\RequirePackage{fix-cm}
\documentclass[twocolumn]{svjour3}          
\smartqed  
\usepackage{amsmath,amssymb,amsfonts}
\usepackage{graphicx}
\usepackage{xcolor}
\usepackage{url}
\usepackage{subfigure}
\usepackage{tcolorbox}
\usepackage{multirow}
\usepackage{listings}
\usepackage{amssymb,amsfonts}
\usepackage{pifont}
\usepackage{xcolor,colortbl} 
\usepackage{listings}
\usepackage{color,hyperref}
\usepackage{marvosym}
\usepackage{float}
\usepackage{pdflscape}

\lstdefinestyle{JavaStyle} {
  backgroundcolor=\color{white},   
  commentstyle=\color{mygreen}, 
  breakatwhitespace=false,
  keywordstyle=\color{violet},
  language=Java,
  stringstyle=\color{blue},
  basicstyle=\scriptsize\ttfamily,
  showstringspaces=false }

\definecolor{mygreen}{rgb}{0,0.6,0}
\definecolor{mygray}{rgb}{0.95,0.95,0.95}
\definecolor{myred}{rgb}{0.5,0,0}

\def\thickhline{\noalign{\hrule height1.3pt}}

\usepackage{xspace}
\newcommand*{\ie}{i.e.,\@\xspace}
\newcommand*{\eg}{e.g.,\@\xspace}

\makeatletter
\newcommand*{\etc}{%
	\@ifnextchar{.}%
	{etc}%
	{etc.\@\xspace}%
}
\makeatother
\newcommand*{\etal}{\emph{et~al.}\@\xspace}
\newcommand\revised[1]{\textcolor{black}{#1}}

\definecolor{Crimson}{rgb}{0.86,0.07,0.23}
\definecolor{Gold}{rgb}{0.98,0.65,0.0}

\definecolor{darkgray}{gray}{0.78}
\definecolor{lightgray}{gray}{0.85}
\definecolor{verylightgray}{gray}{0.95}

\begin{document}

\title{SGD method for entropy error function with smoothing $l_0$ regularization for neural networks}


\author{Trong-Tuan Nguyen  \and  Van-Dat Thang \and Nguyen Van Thin$^{*}$ \and Phuong T. Nguyen}

\institute{Trong-Tuan Nguyen \at
VNPT AI, Vietnam 
	\email{\href{mailto:tuannt99@vnpt.vn}{tuannt99@vnpt.vn}}
\and 
	Van-Dat Thang \at
	Viettel High Technology Industries Corporation, Hanoi, Vietnam
	\email{\href{mailto:datthangvan@gmail.com}{datthangvan@gmail.com}}
\and
 \Letter~Nguyen Van Thin
 \at
	Department of Mathematics, Thai Nguyen University of Education, Luong Ngoc Quyen Street, Thai Nguyen, Vietnam 
	\email{\href{mailto:thinnv@tnue.edu.vn}{thinnv@tnue.edu.vn}}
	\and
	Phuong T. Nguyen \at
	Department of Information Engineering, Computer Science and Mathematics, Universit\`a degli studi dell'Aquila, Italy \\ 
	\email{\href{mailto:phuong.nguyen@univaq.it}{phuong.nguyen@univaq.it}}    
 }

\maketitle

\begin{abstract}
The entropy error function has been widely used in neural networks. Nevertheless, the network training based on this error function generally leads to a slow convergence rate, and can easily be trapped in a local minimum or even with the incorrect saturation problem in practice. In fact, there are many results based on entropy error function in neural network and its applications. However, the theory of such an algorithm and its convergence have not been fully studied so far. To tackle the issue, this works proposes a novel entropy function with smoothing $l_0$ regularization for feed-forward neural networks. An empirical evaluation has been conducted on real-world datasets to demonstrate that the newly conceived algorithm allows us to substantially improve the prediction performance of the considered neural networks. More importantly, the experimental results also show that the proposed function brings in more precise classifications, compared to well-founded baselines. The work is novel as it enables neural networks to learn effectively, producing more accurate predictions compared to state-of-the-art algorithms. In this respect, it is expected that the 
algorithm will contribute to existing studies in the field, advancing research in Machine Learning and Deep Learning. 
\revised{}
\end{abstract}

\section{Introduction}
\def\theequation{1.\arabic{equation}}
\setcounter{equation}{0}

The gradient descent method~\cite{SHARMA2021107084} is one of the most powerful local optimization algorithms. In the last decades, multilayer feed-forward neural networks have been widely applied across various application domains, such as classification, computer vision, natural language processing, to name a few~\cite{10.1016/j.amc.2017.05.010,SENHAJI20201}. In neural networks, the square error function is usually used and many results on the convergence and stability of gradient method on that function are established
	~\cite{GULIYEV2018262,XIAO2021173}.

 In fact, a network with a compact structure is in favor since it has stronger generalization ability compared to a network with much more complex structures, given that both networks reach the same prediction performance \cite{10.1007/s00521-014-1730-x}. However, the network training based on square error function can lead to a slower convergence rate and easily trap in a local minimum or even with the incorrect saturation problem in practice. To address this issue, 
		different error functions have been conceptualized, \eg the entropy error function, Hinge loss~\cite{NIPS1998_a14ac55a}, Huber loss~\cite{hastie01statisticallearning}. The entropy error function was proposed by Karayiann \cite{160170} and then modified by Oh \cite{10.1016/j.neucom.2010.11.024}. Empirical studies have shown that the gradient method based on the entropy error function performs well in convergence and stability. There are many results based on entropy error function in neural network and its applications \cite{8844264,10.1007/s00521-018-3933-z}. However, the theory of such an algorithm and its convergence have not been fully studied so far. 	
		Recently, the batch gradient method with smoothing $l_0$ regularization for feed-forward neural networks was studied by other authors such as Zhang \etal \cite{10.1007/s00521-014-1730-x}. Afterward, Xiong \etal \cite{xiong_convergence_2020} proposed a new algorithm based on entropy error function for feed-forward neural networks. 
		

		\subsection{Problem definition}
		
		
		$l_p$ regularization learning is a popular training method used in neural networks. It is a special form of the $h$-regularization method.
		General speaking, it 
		is to modify the loss function with the following form
		\begin{align}\label{new2}
			{\bf E({\bf w})=I({\bf w})+\lambda h({\bf w}),}
		\end{align}
		where $\lambda>0$ is the penalty factor, $I({\bf w})$ is the original loss function and $h({\bf w})$ is the regularized part. It is worth noting that $I$ and $h$ are nonnegative functions. If 
		$h({\bf w})=||{\bf w}||_t^{t},$ is chosen where $||{\bf w}||_t^{t}=|z_1|^{t}+\dots+|z_p|^{t}$ with ${\bf w}=(z_1,\dots,z_p)$ and
		$t=2,1,\dfrac{1}{2},$ then we get $l_2,$ $l_1,$ and $l_{1/2}$ regularization respectively. There exists a line of work related to $l_2$ regularization~\cite{LH,XZ}, $l_1$ regularization \cite{DBLP:journals/nn/Ishikawa96,DBLP:journals/neco/Williams95}, or $l_{1/2}$ regularization~\cite{DBLP:journals/ijon/LiZW18,DBLP:journals/ijon/LiuWFYW14}.
		
		
		The observation see that $l_p$ regularization methods tend to bear more sparse results as $t\to 0.$ Then it 
		is interesting to investigate $l_p$ regularization as $p\to 0.$ However, 
		$l_0$ regularization is an NP-hard problem \cite{1542412,Wang2013} and it is not possible to directly use gradient method due to the non-continuous characteristic of the $l_0$ norm. Malek-Mohammadi \etal~\cite{7500117} studied the sparsity approximation for compressed sensing. A general formulation is to find $\tilde {\bf x}$ from the following equation	
		\begin{align}\label{m0}
			{\bf b}={\bf A}\tilde{\bf x}+{\bf w},
		\end{align}
		
		in which ${\bf b}\in \mathbb R^n$ is the vector of measurements, ${\bf A}\in \mathbb R^{n\times m}\;(n<m)$ is the sensing matrix, 
		$\tilde {\bf x}\in\mathbb R^m$ is the unknown sparse vector with $s\ll m$ nonzero elements, and ${\bf w}$ is the probable additive noise.
		The aim is to accurately estimate $\tilde {\bf x}.$ When (\ref{m0}) is underdetermined, recovery of $\tilde {\bf x}$ from ${\bf b}$ and 
		${\bf A}$ is ill-posed unless there is a priori that $\tilde {\bf x}$ resides in a low-dimensional space. Then there is 
		the approximation problem	
		\begin{align}\label{m01}
			\min_{\bf x}||{\bf x}||_0\;\text{subject to}\; ||{\bf A}{\bf x}-{\bf b}||\le \varepsilon,
		\end{align}
		
		where $||{\bf x}||_0$ is $l_0$ norm of ${\bf x}$ defined as its number of nonzero entires and $||.||$ stands for $l_2$ norm, and $\varepsilon\ge ||{\bf w}||$
		is a constant. In the $l_0$ regularization, the general method is approximated $l_0$ norm by 
		smoothing function, as done by various studies \cite{yang_design_2020,10.1007/s00521-014-1730-x}. %

		\subsection{State-of-the-art research}
		
		
		Moulay \etal \cite{MOULAY201929} 
		study the properties of the sign gradient descent algorithms involving the sign of the gradient instead of the gradient itself and first introduced in the RPROP algorithm. The authors presented two outcomes of convergence for local optimization, \ie nominal systems without uncertainty and 
		with uncertainties. New sign gradient descent algorithms including the dichotomy algorithm DICHO are applied on several examples to show their effectiveness in terms of speed of convergence.
		
		Starting from the observation that the regular SGD algorithm tends to be stuck in a local optimum, Li \etal \cite{LI2022196} proposed an Adjusted Stochastic Gradient Descent (ASGD) for Latent Factor Analysis. The authors implemented an adjustment mechanism, taking into consideration the bi-polar gradient directions during optimization. 
		Moreover, the model's hyper-parameters are implemented in a self-adaptive manner by means of the particle swarm optimization (PSO) algorithm. The empirical evaluation demonstrated that the proposed approach obtains a better performance compared to various state-of-the-art studies. 
		

		Senhaji \etal \cite{SENHAJI20201} conceived a novel approach to the optimization of the Multilayer Perceptron Neural Network to deal with the generalization problem. The method led to a set of solutions called Pareto front, being the optimal solutions set, the adequate MLPNN need to be extracted. The authors demonstrated that the proposed approach can decrease the neural networks topology and enhance generalization performance, in addition to a good classification rate compared to different methods.
		
		Extended extreme learning machine (ELM) was proposed by Han \etal~\cite{HAN2014128} as a means for training the weights of a class of hierarchical feed-forward neural network (HFNN). Different from conventional single-hidden-layer feed-forward networks (SLFNs), hierarchical ELM (HELM) has been developed on top of the hierarchical structure which is able to learn sequential information online, and one may simply choose hidden layers and then only need to adjust the output weights linking the hidden layer and the output layer. Through an empirical evaluation, the authors showed that the proposed HELM approach is effective. 
		
		Zhang \etal \cite{10.1007/s00521-014-1730-x} proposed a batch gradient method with smoothing $l_0$ regularization for training the feed-forward neural networks, by approximating the $l_0$ with smoothing functions. By means of an evaluation, the authors proved that the proposed approach contributes to sparse results, and thus helping to prune the network efficiently.

		Egwu \etal \cite{DBLP:journals/apin/EgwuMS23} proposed a novel technique for continuous input selection through the utilization of a dimensional reduction approach involving  `structured'  L2-norm regularization. The technique involves identifying the most informative feature subsets from a given dataset using an adaptive training mechanism. To address the non-differentiability issue associated with the gradient of the structured norm penalty, they introduced a modified gradient approach during training. When applied to process datasets, their method demonstrates that structured L2-norm penalization can effectively select the most informative inputs for artificial neural networks.

		A series of studies have been conducted to solve various issues in Machine Learning with respect to error entropy. Yang \etal~\cite{10261_336924,YANG2023126240,e24040455} introduced a spike-based framework with minimum error entropy, named MeMEE, using the entropy theory to establish the gradient-based online meta-learning scheme in a recurrent SNN architecture. Similar work~\cite{10325627,10235316,10310025} has been done to solve the problem of robust and energy-efficient learning for spike-based algorithms in information bottleneck framework. For long short-term memory (LSTM) neural networks, research~\cite{WANG2023128677} has been done to propose an improved robust multi-time scale singular filtering-Gaussian process regression-long short-term memory modeling method for the estimation of remaining capacity, and prediction of life of lithium-ion batteries~\cite{DBLP:journals/ress/WangFJTF23}.

		The most relevant work to ours is Xiong \etal \cite{xiong_convergence_2020}, where the authors presented an entropy error function for feed-forward neural networks. The weak and strong convergence analysis of the gradient method based on the entropy error function with batch input training patterns is strictly proved. 
		The authors investigated if the proposed function is effectiveness, using a series of numerical examples. 
		Compared with the square error function, the proposed one enables faster learning as well as better generalization. The experimental results show that the proposed approach obtains a better performance by almost all of the considered datasets. 
		
		Compared to existing studies, the work is novel since it allows feed-forward neural networks to substantially improve their performance, increasing effectiveness. We anticipate that the application of the proposed work will introduce more precise predictions for neural networks.
		

		\subsection{Contributions}
		
		In this paper, \emph{a novel algorithm for entropy error function with smoothing $l_0$ regularization is conceived}. The ultimate aim is to develop a new entropy error function with L0-regularization, and introduce the Gradient descent training method for this class function. In general, this class loss function is a special case of loss function with non-convex, and the convergence of SGD method in non-convex setting is established with the h-regularization.
		Empirical experiments have been performed on real-world datasets to validate the proposed approach. The experimental results show that the algorithm brings more accurate prediction as well as converges faster. 
		In this respect, the paper makes the following contributions:
		
		\begin{itemize}	
			\item The work presents a novel gradient descent algorithm based on the entropy error function with smoothing $l_0$ regularization for feed-forward neural network (EEGML0). It is a special form of gradient descent method with $h$-regularization for nonconvex function; 
			\item The convergent properties of \emph{EEGML0} are solved under the 
			conditions for strong convergence;
			\item An empirical evaluation on real-world datasets to validate the effectiveness of the proposed algorithm;
			\item The source code is available online to faclitate future research.\footnote{\url{https://github.com/santapo/EEGML0}} 
		\end{itemize}	

	The paper is organized as follows. Section~\ref{sec:Gradient} introduces a gradient descent method on the entropy error function with smoothing $l_0$ regularization. This section also studies the convergence of the proposed method. In Section~\ref{sec:Proof}, the mathematical proof for the proposed algorithm is presented in detail. 
	Afterward, Section~\ref{sec:Evaluation} reports and analyzes the numerical examples to study the performance of \emph{EEGML0}, as wel as to compare it with well-established baselines. 
	Finally, Section~\ref{sec:Conclusions} sketches future work, and concludes the paper.

	\def\theequation{3.\arabic{equation}}
	\setcounter{equation}{0}
	
	\section{The proposed SGD method} \label{sec:Gradient}
	This section presents in detail a novel algorithm for entropy error function with with smoothing $l_0$ regularization. Before stating the result, we recall some notations as follows. Assume that the number of neurons for the input and output layers are $J$ and $1,$ respectively. Denote the weighted 
	vector by $W=(w_1, w_2, \dots, w_p)^{T}$ and $g(x)=\dfrac{1}{1+e^{-x}}$ is the sigmoid function~\cite{DBLP:journals/nn/MinaiW93}. Then we have $g'(x)=(1-g(x))g(x)$ for all 
	$x\in\mathbb R$ and 
	\begin{align}\label{a0}
		0\le g'(x)\le \dfrac{1}{4}.
	\end{align} 
	Let $\{\xi^{j},\zeta^{j}\}_{j=1}^{J}\subset \mathbb R^{p}\times\mathbb R$ be the input training 
	samples. For the given weight vector $W,$
	the output of the neural network is given by
	\begin{align}\label{ct1}
		y^{j}=g(W.\xi^{j}).
	\end{align}
	
	Let 
	\begin{align}\label{ct2}
		E(W)&=-\sum_{j=1}^{J}[\zeta^{j}ln y^{j}+(1-\zeta^{j})ln(1-y^{j})]\notag\\
		&=-\sum_{j=1}^{J}[\zeta^{j}ln(g(W.\xi^{j}))+(1-\zeta^{j})ln(1-g(W.\xi^{j}))].
	\end{align}
	be the entropy error function. It is assumed that $\zeta^{j}\in [0,1]$ for all $j=1,\dots,m.$ Xiong \etal \cite{xiong_convergence_2020} defined the gradient descent training method for this error function. Recently, some authors studied the error functions with $l_t$ regularization $$ L(W)=I(W)+\lambda||W||_{p}^{p},$$
	where $I$ is a loss function associated with training samples, $t=2,1$ and $\dfrac{1}{2},$ and $||z||_t^{t}=(|z_1|^{t}+\dots+|z_p|^{t})^{1/t}$ for 
	$z=(z_1,\dots,z_p)^{T}.$ The best $l_t$ regularizer satisfies the sparsity requirement is the $l_0$ regularizer defined by 
	$$ ||z||_0^{0}=\lim_{t\to 0}((|z_1|^{t}+\dots+|z_p|^{t})^{1/t})^{t}=\lim_{t\to 0}(|z_1|^{t}+\dots+|z_p|^{t}).$$ 
	Then $||z||_0^{0}$ is equal to the number of nonzero entries of the vector $z.$ In this paper, the new error function is considered with the form
	\begin{align}\label{ct3}
		L(W)=E(W)+\lambda ||W||_0^{0},
	\end{align}
	where $\lambda>0$ is the regularization coefficients to balance the trade off between the training accuracy and the network complexity. In (\ref{ct3}), 
	the $l_0$ norm is not a continuous function, then the minimizing of (\ref{ct3}) is an NP-hard problem in combining the optimization method. To overcome this difficulty, the continuous function of vector variable is used:
	\begin{align}\label{a4}
		H_{\sigma}(z)=\sum_{i=1}^{p}h_{\sigma}(z_i)
	\end{align}
	to approximate the $l_0$ regularizer, where $h_{\sigma}(.)$ is continuously differentiable on $\mathbb R$ and satisfies the following condition 
	\begin{align}\label{a5}
		\lim_{\sigma\to 0}h_{\sigma}(t)=
		\begin{cases}
			&1 \;\; \text{if}\;\; t\ne 0\\
			&0\;\; \text{if}\;\; t=0
		\end{cases}.
	\end{align}
	Here, $\sigma$ is a positive number used to control the degree how $H_{\sigma}(z)$ approximates the $l_0$ regularizer. From (\ref{a4}) and (\ref{a5}),	it is known that $H_{\sigma}(z)$ approximates the number of nonzero entries of $z$ (\ie the $l_0$ regularier) as $\sigma\to 0.$There is some good choice
	of $h_{\sigma}(t)$ such as $1-\text{exp}(-\dfrac{t^2}{2\sigma^2}), 1-\dfrac{\sigma^2}{t^2+\sigma^2},1-\dfrac{sin(t/\sigma)}{t/\sigma}$ and~\cite{DBLP:journals/jscic/GuoLQY21}
	$$\dfrac{\text{exp}(\dfrac{t^2}{2\sigma})-\text{exp}(-\dfrac{t^2}{2\sigma})}{\text{exp}(\dfrac{t^2}{2\sigma})
		+\text{exp}(-\dfrac{t^2}{2\sigma})}.$$
	Hence, the error function $L(W)$ in (\ref{ct3}) is modified as follows:
	\begin{align}\label{a6}
		L(W)=E(W)+\lambda H_{\sigma}(W).
	\end{align}
	Denote $L_W(W)=(L_{w_1}(W),\dots,L_{w_p}(W))$ by the gradient of $L$ with respect to $W.$ Starting from an arbitrary initial value $W^{0},$ the batch gradient method with the smoothing regularization term updates the weights $\{W^n\}$
	iteratively by 
	\begin{align}\label{a7}
		W^{m+1}=W^{m}-\eta L_W(W^m), m=0,1,\dots,
	\end{align}
	where $\eta>0$ is the learning rate. There are the following assumptions:
	
	\begin{itemize}
		\item $(A_1)$ For any fixed positive parameter $\sigma\in \mathbb R,$ $h_{\sigma}$ is twice differentiable and uniformly bounded for both 
		the first and the second derivatives on $\mathbb R.$
		\item $(A_2)$ There exists a closed bounded region $\Omega\subset \mathbb R^p$ such that $\{W^{m}\}\subset\Omega$ and the set $\Omega_0=\{W\in \Omega: L_W(W)=0\}$ contains finite points.
	\end{itemize}
	
	The result is stated as follows:

	\begin{theorem}\label{th1}
		Let $L(W)$ be the error function given by (\ref{a6}) and the sequence $\{W^{m}\}$ is generated by the algorithmic (\ref{a7}). If the learning rate 
		$\eta\in (0, \dfrac{2}{L}),$
		where $L>0$ is defined by (\ref{m1}) and the assumptions $(A_1)$ hold, then there exists $\beta>0$ such that
		\begin{align}\label{a8}
			L(W^{m+1})\le L(W^m)-\beta||L_W(W^m)||^2
		\end{align}
		and 
		\begin{align}\label{a9}
			\lim_{m\to\infty}||L_W(W^m)||=0.
		\end{align}
		Moreover, if the assumption $(A_2)$ holds, then there exists a $W^{*}\in \Omega$ such that the following strong convergence holds 
		\begin{align}\label{a10}
			\lim_{m\to\infty}W^m=W^{*}.
		\end{align}
	\end{theorem}
	
	Now, the proof of Theorem \ref{th1} is given, and there is the following result:
	
	\begin{lemma}\cite{ORTEGA19701,xiong_convergence_2020} \label{lm1} Suppose that $f: \mathbb R^{p}\to \mathbb R$ is a continuous and differentiable on a compact set $\Omega$ and 
		$\Omega_0=\{W\in \Omega: \nabla f(W)=0\}$ has only finite number points. If a sequence $\{W^m\}_{m=0}^{\infty}\subset\Omega$ satisfies 
		$$ \lim_{m\to\infty}||W^{m+1}-W^m||=0,\;\text{and}\; \lim_{m\to\infty}||\nabla f(W^m)||=0,$$
		then there exists a $W^{*}\in \Omega_0$ such that $\lim_{m\to\infty}W^m=W^{*}.$
	\end{lemma}
	\begin{proof}[Proof of Theorem \ref{th1}]
		First, it is to show that $L_W(W)$ satisfies the Lipschitz condition on $\Omega.$ It means that there exists $L>0$ such that
		\begin{align}\label{a11}
			||L_W(W_1)-L_W(W_2)||\le L||W_1-W_2||
		\end{align}
		for all $W_1=(w_{11},w_{12},\dots,w_{1p})$ and $W_2=(w_{21},w_{22},\dots,w_{2p})$ in $\Omega.$ Since $\Omega$ is bounded closed in $\mathbb R^p,$
		then it is a compact set. $L_W(W)$, $h_{\sigma}^{'}$ and $ h_{\sigma}^{''}$  are bounded on $\Omega.$ 
		Therefore
		$$E_{W}(W)=\sum_{j=1}^{J}((g(W.\xi^{j})-\zeta^{j}).\xi^{j})$$
		and then 
		$$L_W(W)=E_W(W)+\lambda \nabla H_{\sigma}(W).$$
		Thus, there is the following result:
		\begin{align}\label{a12}
			||L_W(W_1)-L_W(W_2)||\le ||E_W(W_1)-E_W(W_2)||+ \notag\\
			\lambda ||\nabla H_{\sigma}(W_1)-\nabla H_{\sigma}(W_2)||.
		\end{align}
		Applying Lagrange's Mean Value Theorem, the following outcome is obtained
		\begin{align}\label{a13}
			E_W(W_1)-E_W(W_2)&=\sum_{j=1}^{J}((g(W_1.\xi^{j})-\zeta^{j}).\xi^{j})-\notag\\
			\sum_{j=1}^{J}((g(W_2.\xi^{j})-\zeta^{j}).\xi^{j})\notag\\
			&=\sum_{j=1}^{J}((g(W_1.\xi^{j})-(g(W_2.\xi^{j}))\xi^{j}\notag\\
			&=\sum_{j=1}^{J}[g'(t_j)(W_1-W_2).\xi^{j}].\xi^{j},
		\end{align}
		where $t_j$ is in between $W_1.\xi^{j}$ and $W_2.\xi^{j}.$ From (\ref{a0}) and (\ref{a13}), then 
		\begin{align}\label{a14}
			||E_W(W_1)-E_W(W_2)||\le \dfrac{1}{4}||W_1-W_2||\sum_{j=1}^{J}||\xi^{j}||^{2}.
		\end{align}
		In fact, $\nabla H_{\sigma}(W)=(h_{\sigma}'(w_1),\dots,h_{\sigma}'(w_p)).$ Applying Lagrange's Mean Value Theorem and the bounded of 
		$h_{\sigma}^{''}(w_i),i=1,\dots,p,$ there is the following result: 
		\begin{align}\label{a15}
			&||\nabla H_{\sigma}(W_1)-\nabla H_{\sigma}(W_2)||=\sqrt{\sum_{i=1}^{p}(h_{\sigma}'(w_{1i})-h_{\sigma}'(w_{2i}))^2}\notag\\
			&=\sqrt{\sum_{i=1}^{p}(h_{\sigma}''(l_i)(w_{1i}-w_{2i}))^2}\le \text{sup}_{t\in\mathbb R}|h_{\sigma}''(t)|.||W_1-W_2||,
		\end{align}
		where $l_i$ is between $w_{1i}$ and $w_{2i}, i=1,\dots,p,$ $W_1=(w_{11},\dots,w_{1p}),$  $W_2=(w_{21},\dots,w_{2p}).$ 
		Combining (\ref{a12}), (\ref{a14}) and (\ref{a15}) gives the following outcome:
		\begin{align}\label{a16}
			||L_W(W_1)-L_W(W_2)||\le L||W_1-W_2||,
		\end{align}
		where 
		\begin{align}\label{m1}
			L=\dfrac{1}{4}\sum_{j=1}^{J}||\xi^{j}||^{2}+\lambda \text{sup}_{t\in\mathbb R}|h_{\sigma}''(t)|.
		\end{align} 
		Following Lemma 1.2.3 proposed by Nesterov~\cite{alma991023077139705251}, there is:
		\begin{align}\label{tm1}
			L(W_1)\le L(W_2)+<L_W(W_2),W_1-W_2>+\dfrac{L}{2}||W_1-W_2||^2\; \notag\\
			\text{for all}\; W_1,W_2\in\mathbb R^p.
		\end{align}
		Since $0<\eta<\dfrac{2}{L},$ then $\beta=\eta(1-\dfrac{L\eta}{2})>0.$ Applying (\ref{tm1}) for $W_1=W^{m+1}$ and $W_2=W^{m},$ gives:
		\begin{align}\label{a19}
			L(W^{m+1})&\le L(W^{m})-\eta(1-\dfrac{L\eta}{2})||L_W(W^m)||^2\notag\\
			&\le L(W^{m-1})-\beta(||L_W(W^m)||^2+||L_W(W^{m-1})||^2)\notag\\
			&\le \dots \le L(W^{0})-\beta\sum_{i=0}^{m}||L_W(W^{i})||^2.
		\end{align}
		It implies that the sequence $\{L(W^{m})\}$ is a decreasing sequence and
		\begin{align*}
			\beta \sum_{i=0}^{m}||L_W(W^{i})||^2\le  L(W^{0})-L(W^{m}).
		\end{align*}
		Take $m\to \infty$ in above inequality and note that $L(W^{i})\ge 0$ for all $i\ge 0,$ then
		\begin{align*}
			\beta \sum_{i=0}^{+\infty}||L_W(W^{i})||^2\le L(W^{0})<+\infty.
		\end{align*}
		Hence $\lim_{m\to\infty}||L_W(W^{m})||=0.$ From the contruction of the sequence $\{W^m\},$ and
		$\lim_{m\to \infty}||W^{m+1}-W^{m}||=\eta\lim_{m\to\infty}||L_W(W^{m})||=0.$ By Lemma \ref{lm1} and the assumption $(A_2),$ there exists 
		an unique $W^{*}\in \Omega_0$ such that $\lim_{m\to\infty}W^{m}=W^{*}.$
	\end{proof}

	\section{Gradient descent method with  $h$-regularization} \label{sec:Proof}
	
	Following Theorem \ref{th1}, this section studies the Gradient descent method for $h$-regularization. Namely, the loss function (\ref{new2}) is considered, and weight sequence is given as follows:
	\begin{align}\label{a7a}
		W^{m+1}=W^{m}-\eta_m E_{\bf w}(W^m), m=0,1,\dots,
	\end{align}
	where $0<\eta_{\min}\le \eta_m\le\eta_{\max}<1.$
	\begin{theorem}\label{th2}
		Assume that the $I$ and $h$ functions have the gradient Lipschitz with constants $L$ and $L_1,$ respectively. Let's denote $\mathcal L=L+\lambda L_1.$ Then for $0<\eta_{\min}\le \eta_{\max}<\dfrac{2}{\mathcal L},$ then we have
		$$\lim_{m\to\infty}||E_{\bf w}(W^m)||=0.$$ 
		Furthermore, if the Assumption $A_2$ holds, then there exists $W^*$ such that $\lim_{m\to\infty}W^m=W^*.$
	\end{theorem}
	\begin{proof}
		From the assumption of Theorem \ref{th2}, it is seen that $E$ has the gradient Lipschitz with constant $L+\lambda L_1.$ Indeed, there are the following formulas:
		\begin{align*}
			||E_{\bf w}(W_1)-E_{\bf w}(W_2||&=||I_{\bf w}(W_1)+\lambda h_{\bf w}(W_1)-I_{\bf w}(W_2) \notag\\
			+\lambda h_{\bf w}(W_2)||&\le ||I_{\bf w}(W_1)-I_{\bf w}(W_2)||+ \\
			& \lambda ||h_{\bf w} \- (W_1)-h_{\bf w}(W_2)||\notag\\
			&=(L+\lambda L_1)||W_1-W_2||
		\end{align*}
		for all $W_1, W_2$ in $\mathbb R^p.$  Following Lemma 1.2.3 proposed by Nesterov~\cite{alma991023077139705251}:
		\begin{align}\label{tm1a}
			E(W_1)\le E(W_2)+<E_{\bf w}(W_2),W_1-W_2>+ \notag\\
			\dfrac{\mathcal L}{2}||W_1-W_2||^2\;\text{for all}\; W_1,W_2\in\mathbb R^p.
		\end{align}
		Since $0<\eta_{\min}\le \eta_{\max}<\dfrac{2}{\mathcal L},$ then $$\beta_m=\eta_m(1-\dfrac{\mathcal L\eta_m}{2})\ge \eta_{\min}(1-\dfrac{\mathcal L\eta_{\max}}{2}):=\beta_{*}>0.$$ Applying (\ref{a7a}) and (\ref{tm1a}) for $W_1=W^{m+1}$ and $W_2=W^{m},$ gives:
		\begin{align}\label{a19a}
			E(W^{m+1})&\le E(W^{m})-\eta_m(1-\dfrac{\mathcal L\eta_m}{2})||E_{\bf w}(W^m)||^2\notag\\
			&\le E(W^{m-1})-\beta_*(||E_{\bf w}(W^m)||^2+ \notag\\
			||E_{\bf w}(W^{m-1})||^2)\notag\\
			&\le \dots \le E(W^{0})-\beta_*\sum_{i=0}^{m}||E_{\bf w}(W^{i})||^2.
		\end{align}
		It implies that the sequence $\{E(W^{m})\}$ is a decreasing sequence and
		\begin{align*}
			\beta_* \sum_{i=0}^{m}||E_{\bf w}(W^{i})||^2\le  E(W^{0})-E(W^{m}).
		\end{align*}
		Take $m\to \infty$ in above inequality and note that $E(W^{i})\ge 0$ for all $i\ge 0,$ there is:
		\begin{align*}
			\beta_* \sum_{i=0}^{+\infty}||E_{\bf w}(W^{i})||^2\le E(W^{0})<+\infty.
		\end{align*}
		Hence $\lim_{m\to\infty}||E_{\bf w}(W^{m})||=0.$ From the contruction of the sequence $\{W^m\},$ 
		$\lim_{m\to \infty}||W^{m+1}-W^{m}||=\eta_m\lim_{m\to\infty}||E_{\bf w}(W^{m})||=0.$ By Lemma \ref{lm1} and the assumption $(A_2),$ there exists 
		an unique $W^{*}\in \Omega_0$ such that $\lim_{m\to\infty}W^{m}=W^{*}.$
	\end{proof}	
	\begin{remark}
		In Theorem \ref{th2}, taking $I$ by Entropy error function and $h$ by $h_{\sigma}(t),$ and $\eta_m=\eta.$ gives Theorem \ref{th1} from Theorem \ref{th2}. Here, there is $L=\dfrac{1}{4}\sum_{j=1}^{J}||\xi^{j}||^{2}$ and $L_1=\text{sup}_{t\in\mathbb R}|h_{\sigma}''(t)|.$
	\end{remark}

	\section{Empirical evaluation} \label{sec:Evaluation}
	
     To assess the performance of the EEGML0 algorithm, a two layers neural network has been implemented, in which the number of neural in the input layer is equal to the number of features of the dataset and the number of neural in the output layer is one. Moreover, the Sigmoid activation function is employed in the network. The algorithm was executed with four regularizer functions which are specified in Table~\ref{tab:Configurations}. 
     
     \begin{table}[h!]
     	\footnotesize
     	\scriptsize
     	\caption{The configurations.} 
     	\centering
     	\begin{tabular}{|p{3.0cm}|p{4.5cm}|} \hline
     		\textbf{Notation} & \textbf{Function $h_{\sigma}(z)$} \\ \hline 
     		\textbf{reg1} & $1-\text{exp}(-\dfrac{t^2}{2\sigma^2})$		\\ \hline	
     		\textbf{reg2} & $h_{\sigma}(z)=1-\dfrac{\sigma^2}{t^2+\sigma^2}$    \\ \hline
     		\textbf{reg3} & $1-\dfrac{sin(t/\sigma)}{t/\sigma}$  \\ \hline
     		\textbf{reg4} & $h_{\sigma}(z)=\dfrac{\text{exp}(\dfrac{t^2}{2\sigma})-\text{exp}(-\dfrac{t^2}{2\sigma})}{\text{exp}(\dfrac{t^2}{2\sigma})
     			+\text{exp}(-\dfrac{t^2}{2\sigma})}$  \\ \hline
     	\end{tabular}		
     	\label{tab:Configurations}
     \end{table}
  
     The proposed method was then compared with the EEGM algorithm \cite{xiong_convergence_2020}--a well-established baseline--on various binary classification datasets, each of them is characterized by varying numbers of features and study domain. The nine datasets used in the evaluation, and comparison with the baseline are shown in Table~\ref{tab:Datasets}.


 \begin{table*}[t!]
	\footnotesize
	\caption{The datasets used for evaluation.} 
	\centering
	\begin{tabular}{|p{3.9cm}|p{2.5cm}|p{3.2cm}|p{2.4cm}|} \hline
		\textbf{Dataset} & \textbf{Attribute types} & \textbf{\# of instances} & \textbf{\# of attributes}  \\ \hline 
		Breast cancer coimbra & Integer  & 116 & 10 \\ \hline
		Spect heart & Categorical  & 267 & 22 \\ \hline
		Somerville happiness survey & Integer  & 208 & 60 \\ \hline
		Divorce predictors & Integer  & 170 & 54 \\ \hline
		Connectionist Bench & Integer  & 208 & 60 \\ \hline		
		PIMA & Integer  & 769  & 54 \\ \hline	
		Ionosphere & Integer  & 351  & 34 \\ \hline	
		GunPoint & Integer  & 250 & 150 \\ \hline	
		Coffee & Integer  & 58 & 286 \\ \hline		
	\end{tabular}		
	\label{tab:Datasets}
\end{table*}

\subsection{Datasets}

Nine datasets were collected from an online archive\footnote{The datasets are available in the following link: \url{https://archive.ics.uci.edu/datasets}} for evaluating the proposed algorithms, and comparing it with baselines. The datasets and their characteristics are shown in Table~\ref{tab:Datasets}. Among others, PIMA and Coffee are the datasets with the largest and lowest number of instances, \ie 769 and 58, respectively.

\begin{table*}[t!]
	\caption{Accuracy for different configurations.}
        \centering
	\resizebox{0.7\textwidth}{!}{\begin{tabular}{|l|c|c|c|c} \hline
			Dataset                                          & \# features      & Algorithm                 & Average test accuracy (\%) \\
			\thickhline
			\multirow{7}{*}{Spect heart}                     & \multirow{7}{*}{22} & EEGML0 + reg 1  &          54.18       \\ \cline{3-4}
			&                   & EEGML0 + reg 2    &          54.62        \\ \cline{3-4}
			&                   & EEGML0 + reg 3    &          \textbf{56.25}        \\ \cline{3-4}
			&                   & EEGML0 + reg 4    &          54.12        \\ \cline{3-4}
			&                   & EEGM              &          53.93        \\ \cline{3-4}
                &                   & SGM L2            &          53.68       \\ \cline{3-4}
                &                   & SGM               &          54.06       \\
			\thickhline
			\multirow{7}{*}{Connectionist bench}             & \multirow{7}{*}{60} & EEGML0 + reg 1  &          \textbf{71.50}        \\ \cline{3-4}
			&                   & EEGML0 + reg 2    &          69.68        \\ \cline{3-4}
			&                   & EEGML0 + reg 3    &          70.71        \\ \cline{3-4}
			&                   & EEGML0 + reg 4    &          70.79        \\ \cline{3-4}
			&                   & EEGM              &          69.68        \\ \cline{3-4}
                &                   & SGM L2           &          70.50       \\ \cline{3-4}
                &                   & SGM               &         70.52        \\     
			\thickhline
			\multirow{7}{*}{Divorce predictors}              & \multirow{7}{*}{54} & EEGML0 + reg 1  &         \textbf{100.00 }       \\ \cline{3-4}
			&                   & EEGML0 + reg 2    &          99.90        \\ \cline{3-4}
			&                   & EEGML0 + reg 3    &          99.99        \\ \cline{3-4}
			&                   & EEGML0 + reg 4    &          99.80        \\ \cline{3-4}
			&                   & EEGM              &          99.80        \\ \cline{3-4}
                &                   & SGM L2            &          99.80       \\ \cline{3-4}
                &                   & SGM               &          99.80        \\
			\thickhline
			\multirow{7}{*}{Breast cancer coimbra}           & \multirow{7}{*}{9} & EEGML0 + reg 1   &          77.14        \\ \cline{3-4}
			&                   & EEGML0 + reg 2    &          77.57        \\ \cline{3-4}
			&                   & EEGML0 + reg 3    &          78.14       \\  \cline{3-4}
			&                   & EEGML0 + reg 4    &          77.14        \\ \cline{3-4}
			&                   & EEGM                      &          77.14        \\ \cline{3-4}
                &                   & SGM L2            &          77.14        \\ \cline{3-4}
                &                   & SGM               &          77.14        \\
			\thickhline
			\multirow{7}{*}{Somerville happiness survey}     & \multirow{7}{*}{6} & EEGML0 + reg 1   &          61.16        \\ \cline{3-4}
			&                   & EEGML0 + reg 2    &          60.69        \\ \cline{3-4}
			&                   & EEGML0 + reg 3    &          \textbf{64.18}        \\ \cline{3-4}
			&                   & EEGML0 + reg 4    &          60.46        \\ \cline{3-4}
			&                   & EEGM                      &          59.88        \\ \cline{3-4}
                &                   & SGM L2            &          60.66        \\ \cline{3-4}
                &                   & SGM               &          60.44        \\
			\thickhline
			\multirow{7}{*}{PIMA}                            & \multirow{7}{*}{54} & EEGML0 + reg 1  &          \textbf{73.18}        \\ \cline{3-4}
			&                   & EEGML0 + reg 2    &          72.16                 \\ \cline{3-4}
			&                   & EEGML0 + reg 3    &          71.90                 \\ \cline{3-4}
			&                   & EEGML0 + reg 4    &          72.05        \\ \cline{3-4}
			&                   & EEGM              &          71.96                 \\ \cline{3-4}
                &                   & SGM L2            &          70.64        \\ \cline{3-4}
                &                   & SGM               &          71.53        \\
			\thickhline
			\multirow{7}{*}{Ionosphere}                      & \multirow{7}{*}{34} & EEGML0 + reg 1  &          91.22        \\ \cline{3-4}
			&                   & EEGML0 + reg 2    &          90.99        \\ \cline{3-4}
			&                   & EEGML0 + reg 3    &          \textbf{91.79}        \\ \cline{3-4}
			&                   & EEGML0 + reg 4    &          91.74        \\ \cline{3-4}
			&                   & EEGM                      &          91.41        \\ \cline{3-4}
                &                   & SGM L2            &          91.46        \\ \cline{3-4}
                &                   & SGM               &          91.27        \\
			\thickhline
			\multirow{7}{*}{Gun Point}                       & \multirow{7}{*}{150} & EEGML0 + reg 1 &         75.36         \\ \cline{3-4}
			&                   & EEGML0 + reg 2    &         75.63         \\ \cline{3-4}
			&                   & EEGML0 + reg 3    &         75.06         \\ \cline{3-4}
			&                   & EEGML0 + reg 4    &         \textbf{75.86}         \\ \cline{3-4}
			&                   & EEGM                      &         75.00         \\ \cline{3-4}
                &                   & SGM L2            &          75.59        \\ \cline{3-4}
                &                   & SGM              &          75.79        \\
			\thickhline
			\multirow{7}{*}{Coffee}                          & \multirow{7}{*}{286} & EEGML0 + reg 1 &         99.65         \\ \cline{3-4} \cline{3-4}
			&                   & EEGML0 + reg 2    &         99.39         \\ \cline{3-4}
			&                   & EEGML0 + reg 3    &         \textbf{99.75}         \\ \cline{3-4}
			&                   & EEGML0 + reg 4    &         98.85         \\ \cline{3-4}
			&                   & EEGM                      &         99.46        \\ \cline{3-4}
                &                   & SGM L2            &         98.92        \\ \cline{3-4}
                &                   & SGM              &          98.21        \\ 			\thickhline

	\end{tabular}}
	\label{tab:comparison_table}    
\end{table*}

\subsection{Metrics}

\begin{figure}[h!]
	\centering
	\includegraphics[width=0.80\linewidth]{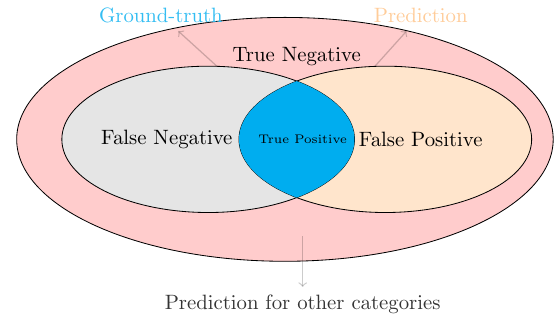}
	\caption{Metric definition~\cite{DBLP:journals/jim/RokachM06}.} 
	\label{fig:MetricsDefinition}
\end{figure}

The definitions of true positive (TP), true negative (TN), false positive (FP), and false negative (FN) are shown in Figure~\ref{fig:MetricsDefinition}.
	
Given a testing sample, the classifier returns a label for it, and the predicted label is compared with the ground-truth one to compute the following scores:

\begin{itemize}
	\item \textit{True positive (TP)}: The items that are correctly classified to the ground-truth category;
	\item \textit{True negative (TN)}: The items that are not classified, and they do not belong to the ground-truth data;
	\item \textit{False positive (FP)}: The classified items which actually do not match the ground-truth data;
	\item \textit{False negative (FN)}: The items that are classified as not correct, but actually they are. 
\end{itemize}		

Then the accuracy metric is defined as follows:

\begin{itemize}
	
%
	
	\item {\bf Accuracy:} The metric is computed as the ratio of correctly classified items to all the returned items:
	\begin{equation} \label{eqn:Accuracy}
		Accuracy = \dfrac{TP + TN}{TP + TN + FP + FN}
	\end{equation}

\end{itemize}




Concerning the hyperparameters, the number of epochs was set to 15,000, and the learning rate to 0.001, while the weight decay is set to 0.0001. To assess the peformance of the EEGML0 algorithm, experiments were conducted using the EEGML0 algorithm with four regularizers, and the EEGM algorithm, and the standard gradient method with and without L2 regularizer, \ie SGM L2 and SGM, respectively on each of the dataset presented in Table~\ref{tab:Datasets}. The EEGML0 algorithm was applied to the datasets with the regularizers to enhance its classification performance. This aims to analyze the effect of EEGML0 with four regularizers on each dataset. To maintain a trade off between consistency and reliability, each dataset is split into training set and validation set following the 70\% : 30\% ratio, if the dataset does not have a separated validation set. Additionally, each experiment was run 20 times, then the scores were averaged out to get the final result. The performance of each algorithm was measured using the average validation accuracy 
on the validation set on 20 runs.

\section{Experimental results}

\begin{figure*}[h!]
	\centering    
	\begin{tabular}{c c}		
		\subfigure[Train loss]{\label{fig:Spect_Heart_train_loss}\includegraphics[width=80mm]{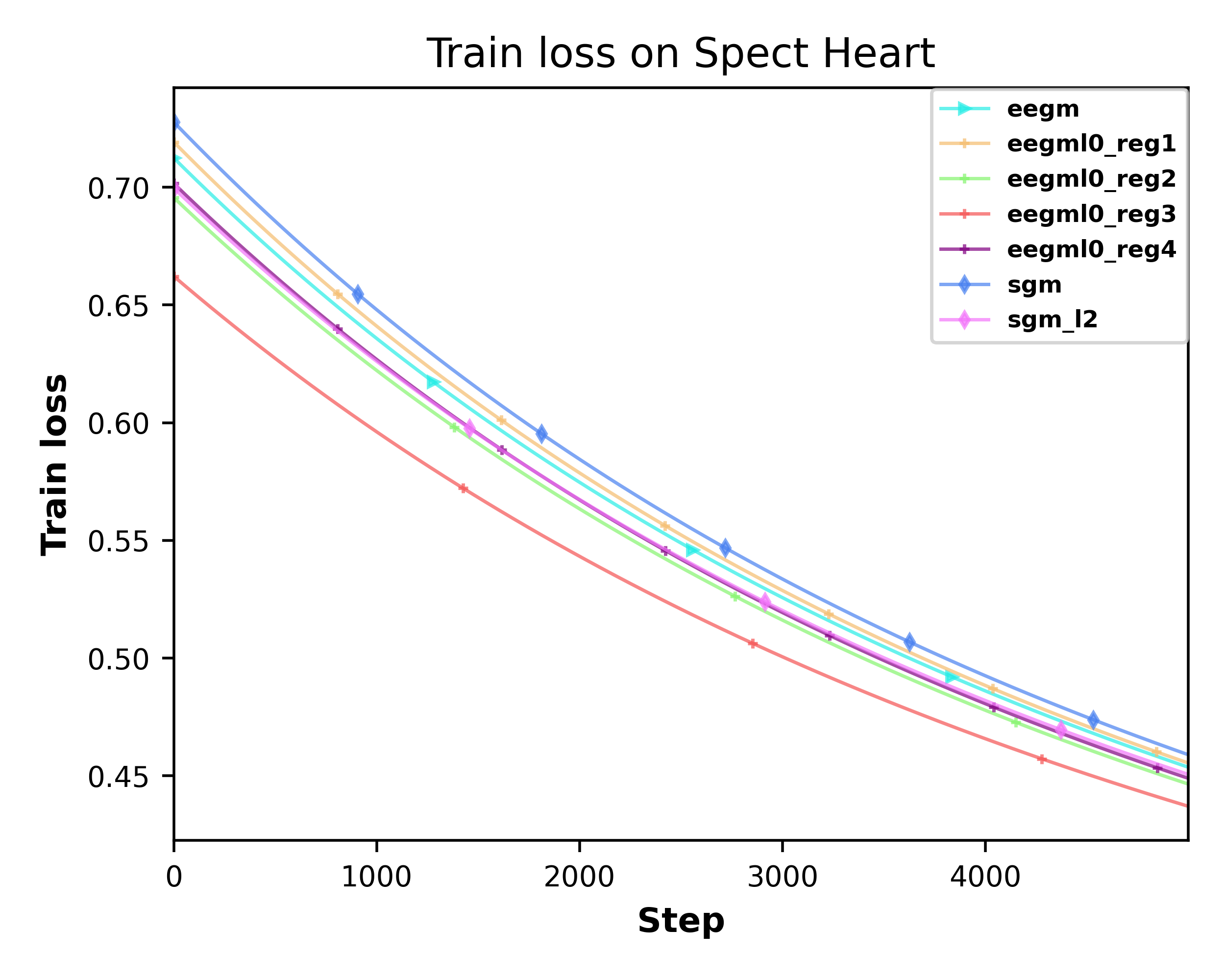}}  &
		\subfigure[Test loss]{\label{fig:Spect_Heart_test_loss}\includegraphics[width=80mm]{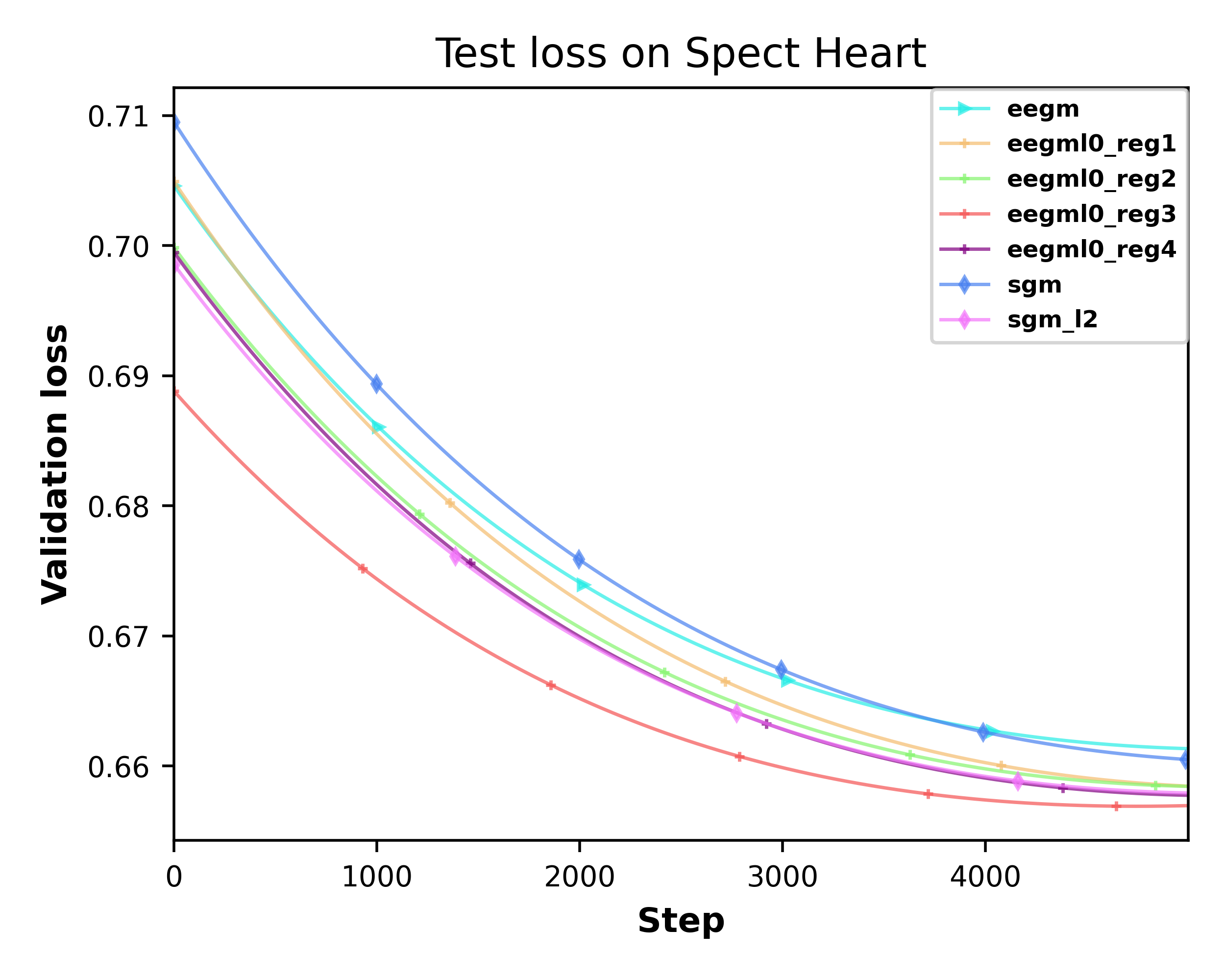}} 
	\end{tabular}
	\vspace{-.2cm}
	\caption{Spect Heart.}
	\label{fig:Spect_Heart}
	\vspace{-.2cm}
\end{figure*}

\begin{figure*}[h!]
	\centering    
	\begin{tabular}{c c}		
		\subfigure[Train loss]{\label{fig:Connectionist_Bench_train_loss}\includegraphics[width=80mm]{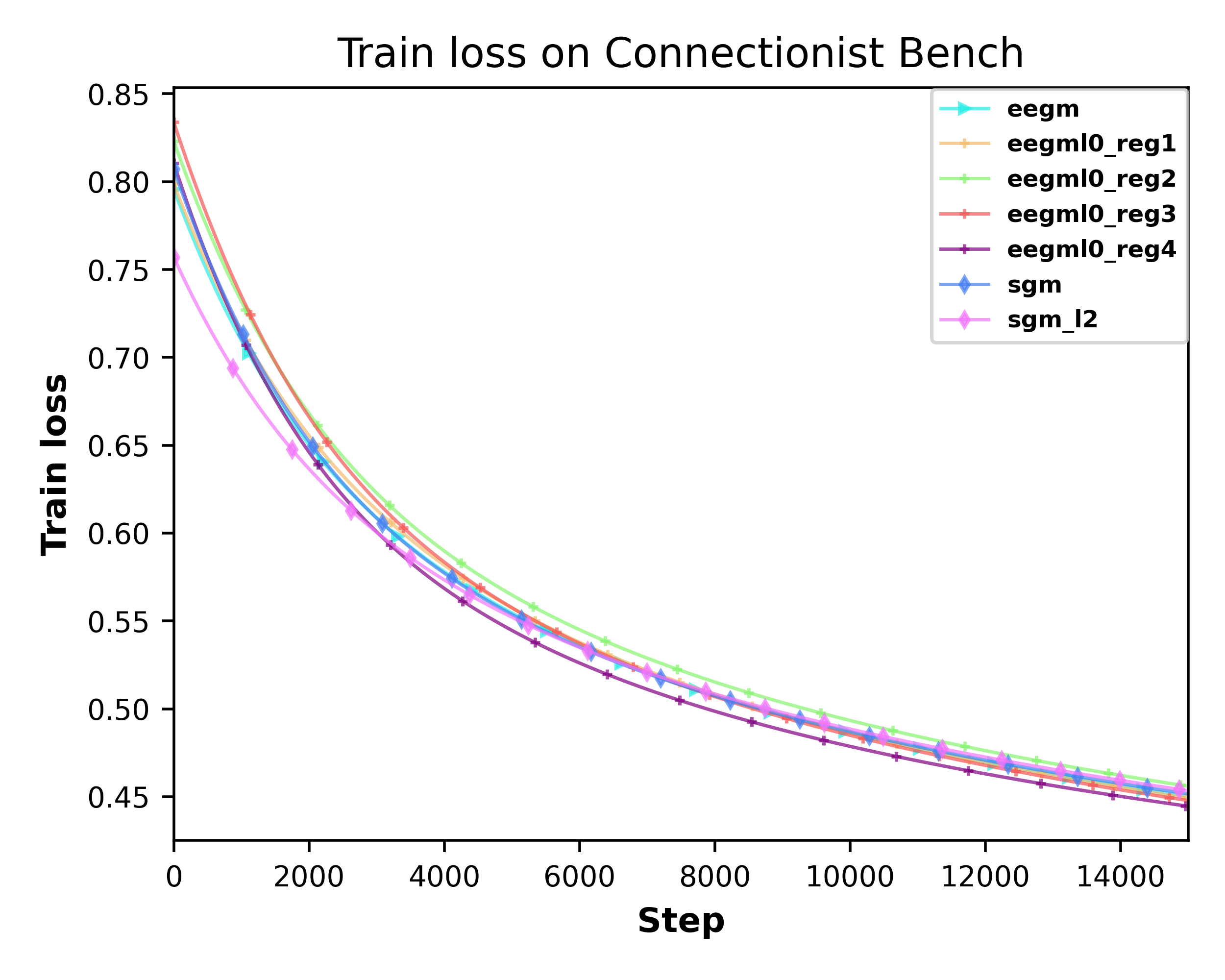}}  &
		\subfigure[Test loss]{\label{fig:Connectionist_Bench_test_loss}\includegraphics[width=80mm]{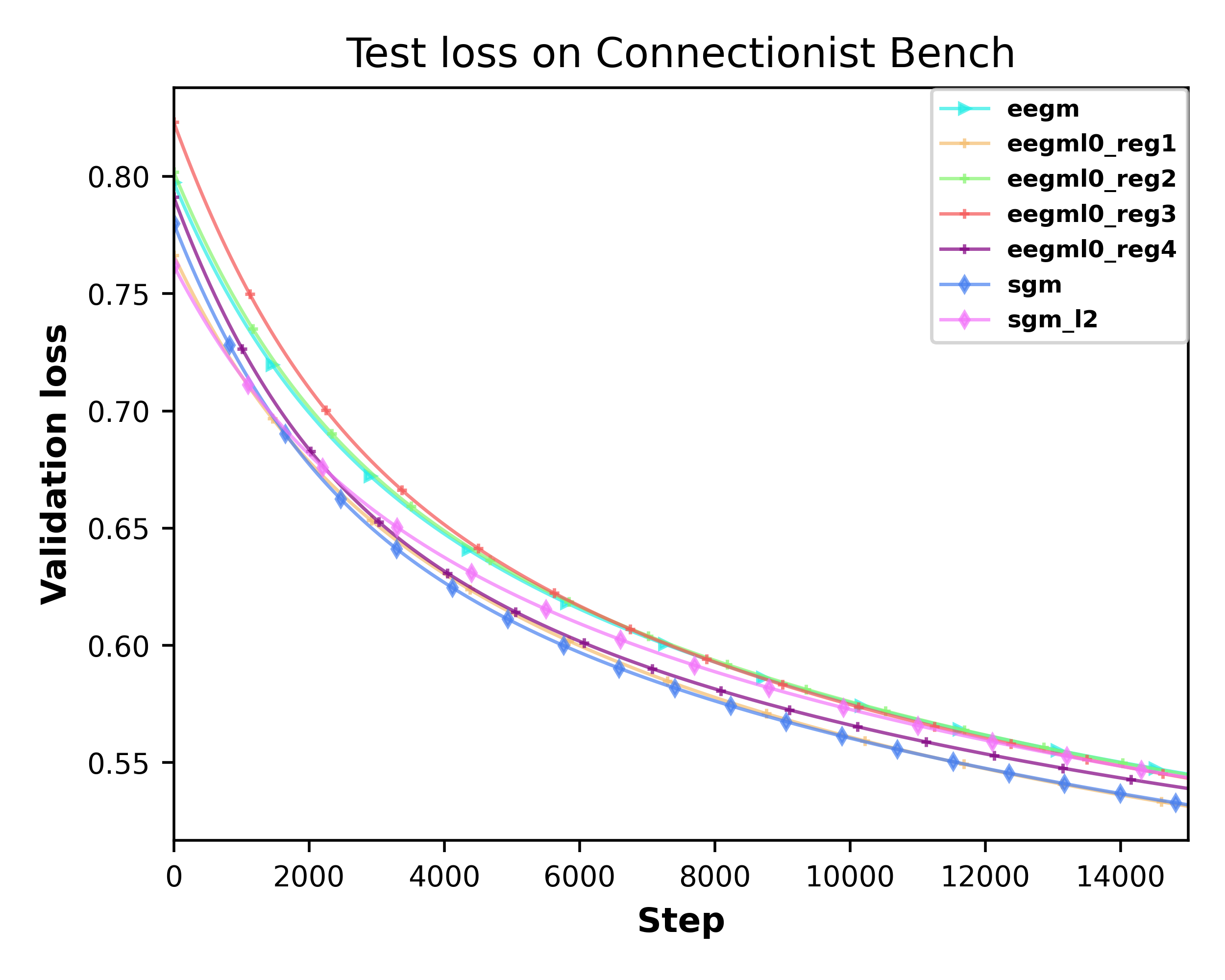}} 
	\end{tabular}
	\vspace{-.2cm}
	\caption{Connectionist Bench.}
	\vspace{-.2cm}
	\label{fig:Connectionist_Bench}
\end{figure*}

\begin{figure*}[h!]
	\centering    
	\begin{tabular}{c c}		
		\subfigure[Train loss]{\label{fig:Divorce_train_loss}\includegraphics[width=80mm]{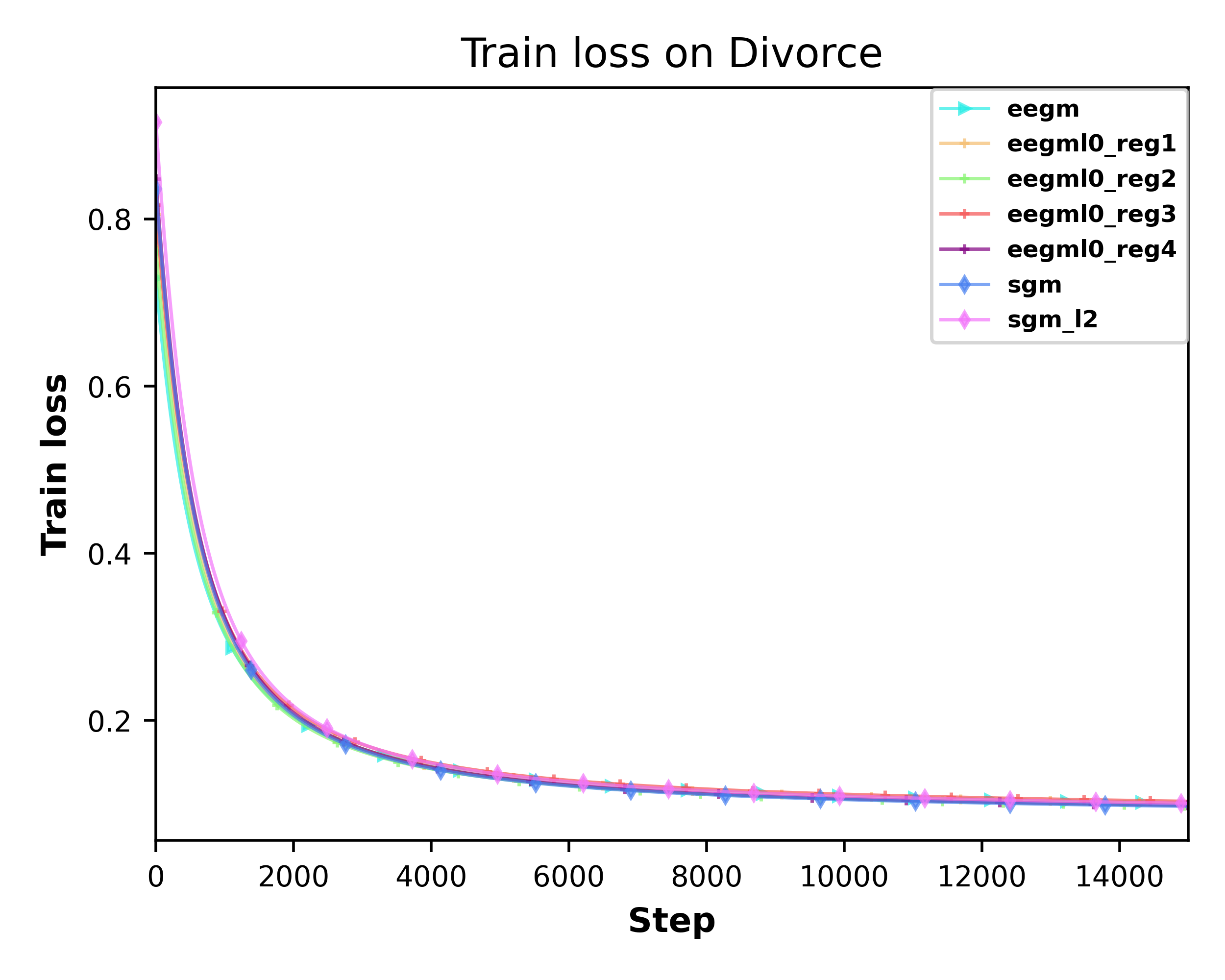}}  &
		\subfigure[Test loss]{\label{fig:Divorce_test_loss}\includegraphics[width=80mm]{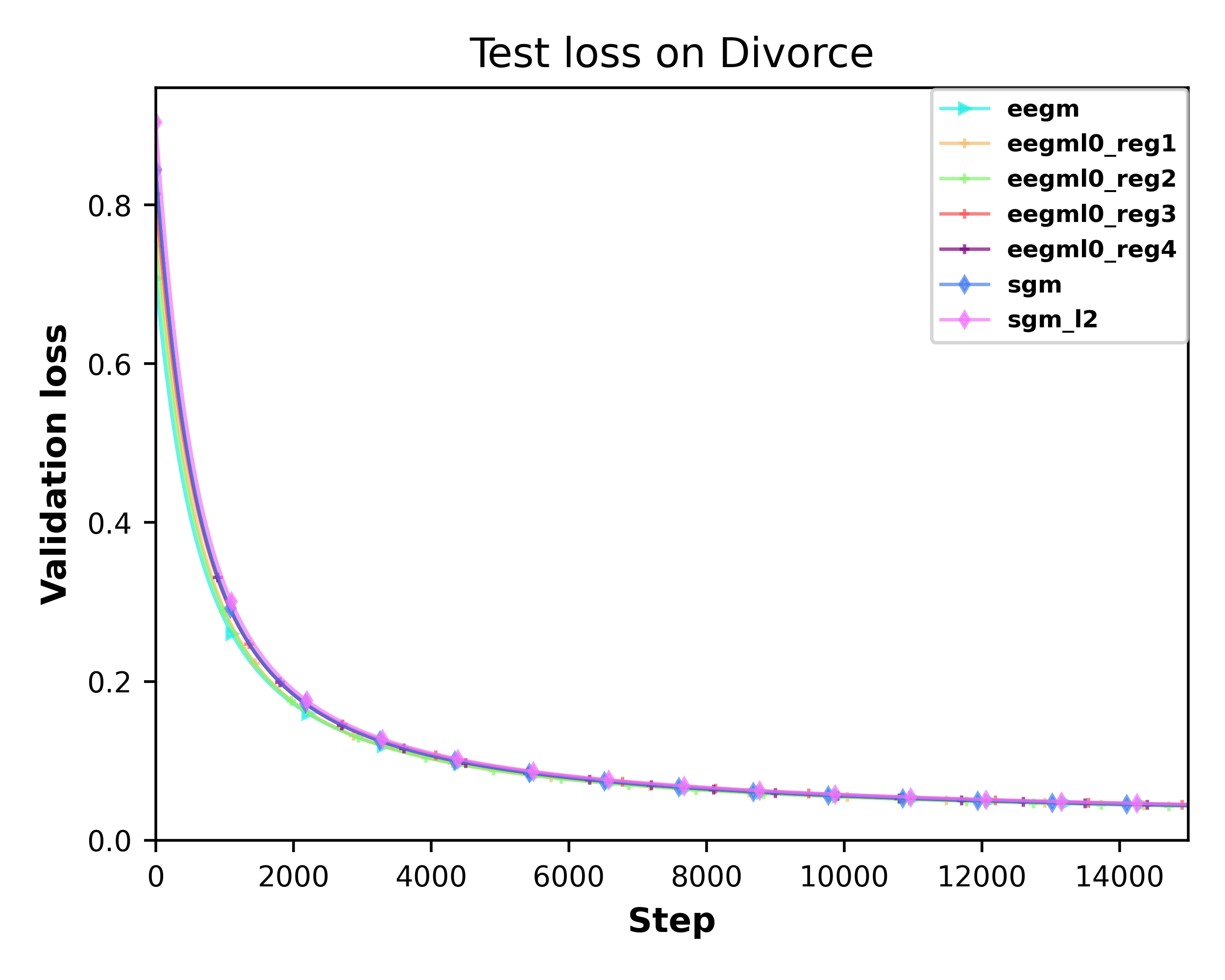}} 
	\end{tabular}
	\vspace{-.2cm}
	\caption{Divorce.}
	\label{fig:Divorce}
	\vspace{-.2cm}
\end{figure*}

\begin{figure*}[h!]
	\centering    
	\begin{tabular}{c c}		
		\subfigure[Train loss]{\label{fig:Breast_Cancer_train_loss}\includegraphics[width=80mm]{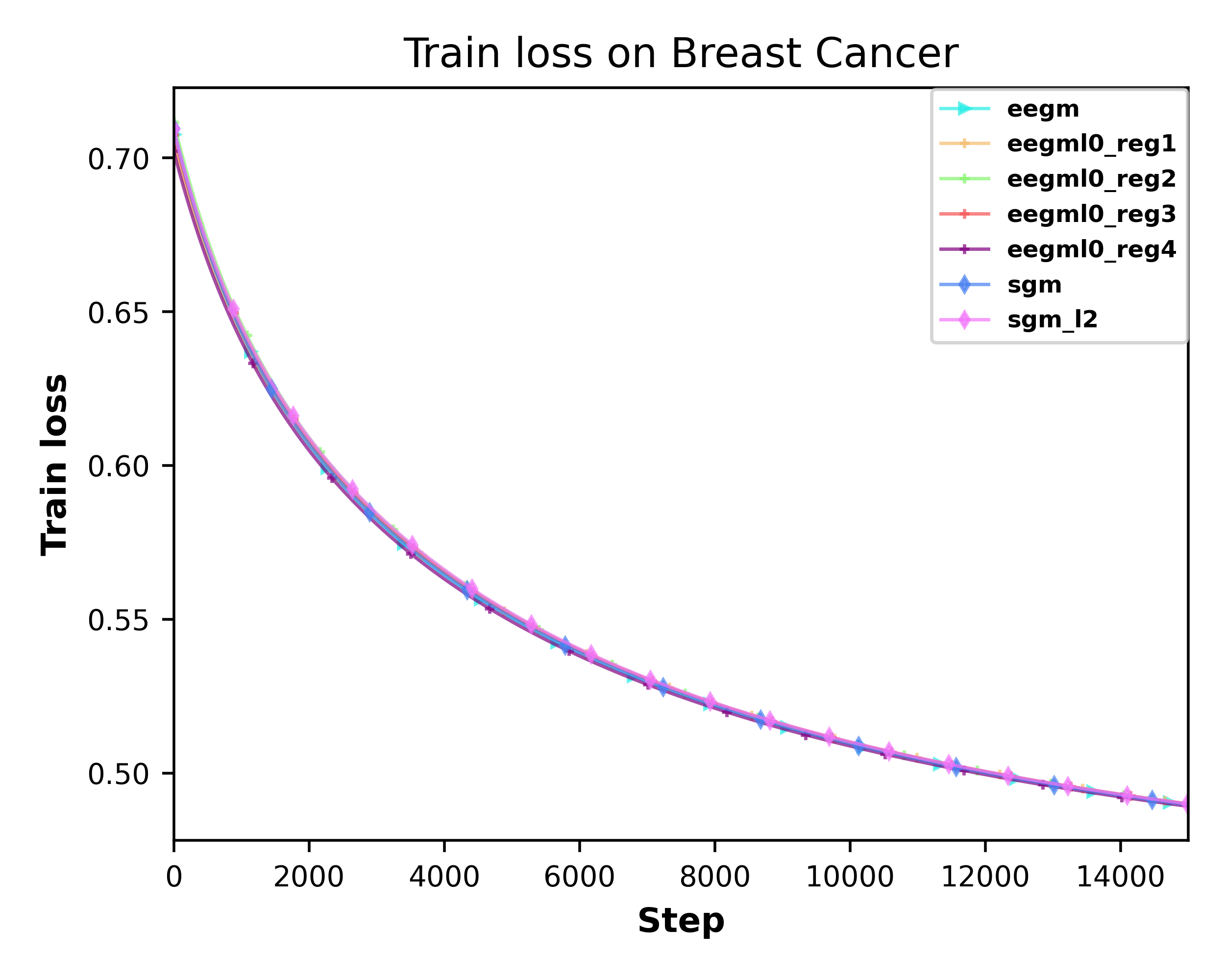}}  &
		\subfigure[Test loss]{\label{fig:Breast_Cancer_test_loss}\includegraphics[width=80mm]{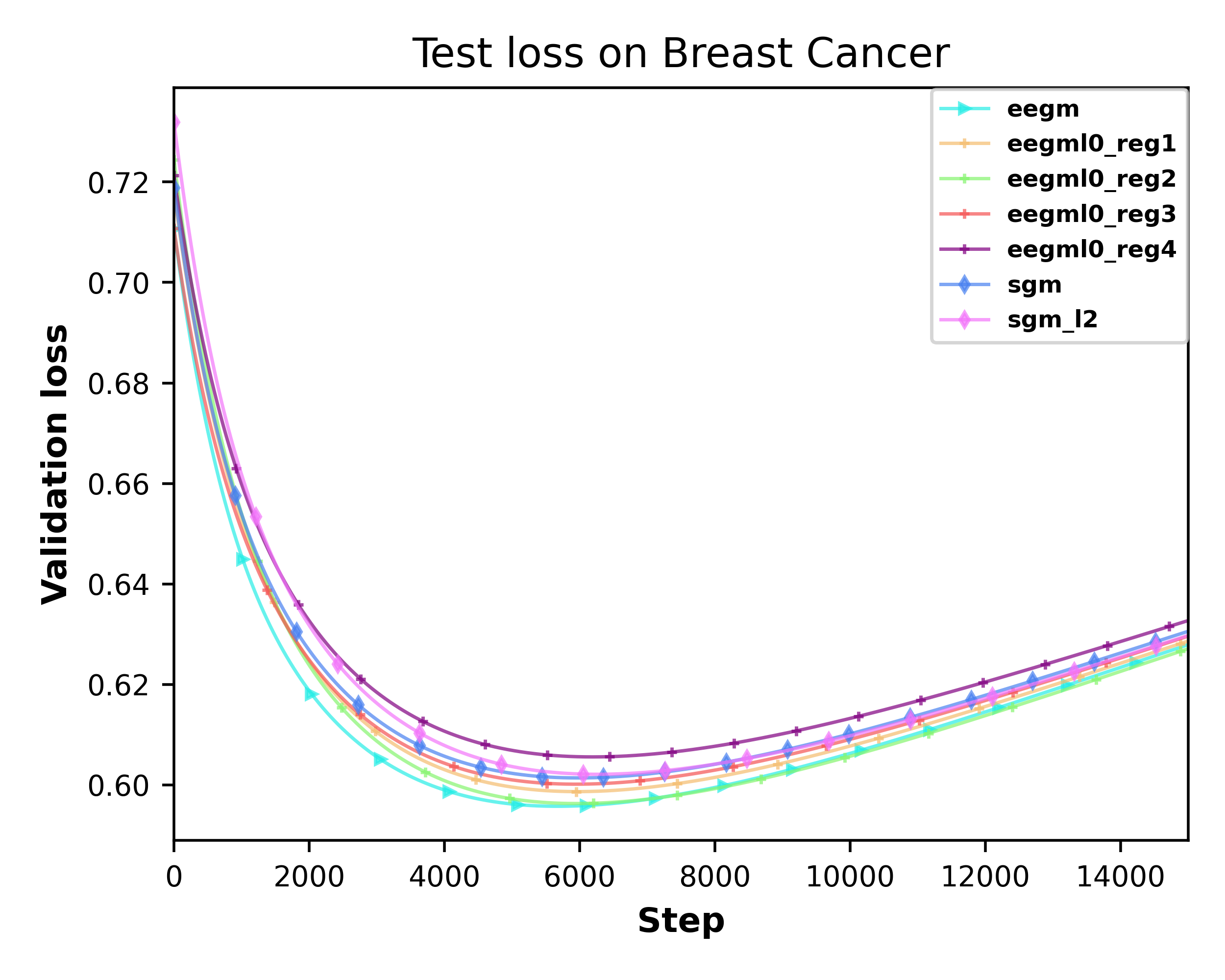}} 
	\end{tabular}
	\vspace{-.2cm}
	\caption{Breast Cancer.}
	\vspace{-.2cm}
	\label{fig:Breast_Cancer}
\end{figure*}

\begin{figure*}[h!]
	\centering    
	\begin{tabular}{c c}		
		\subfigure[Train loss]{\label{fig:Somerville_train_loss}\includegraphics[width=80mm]{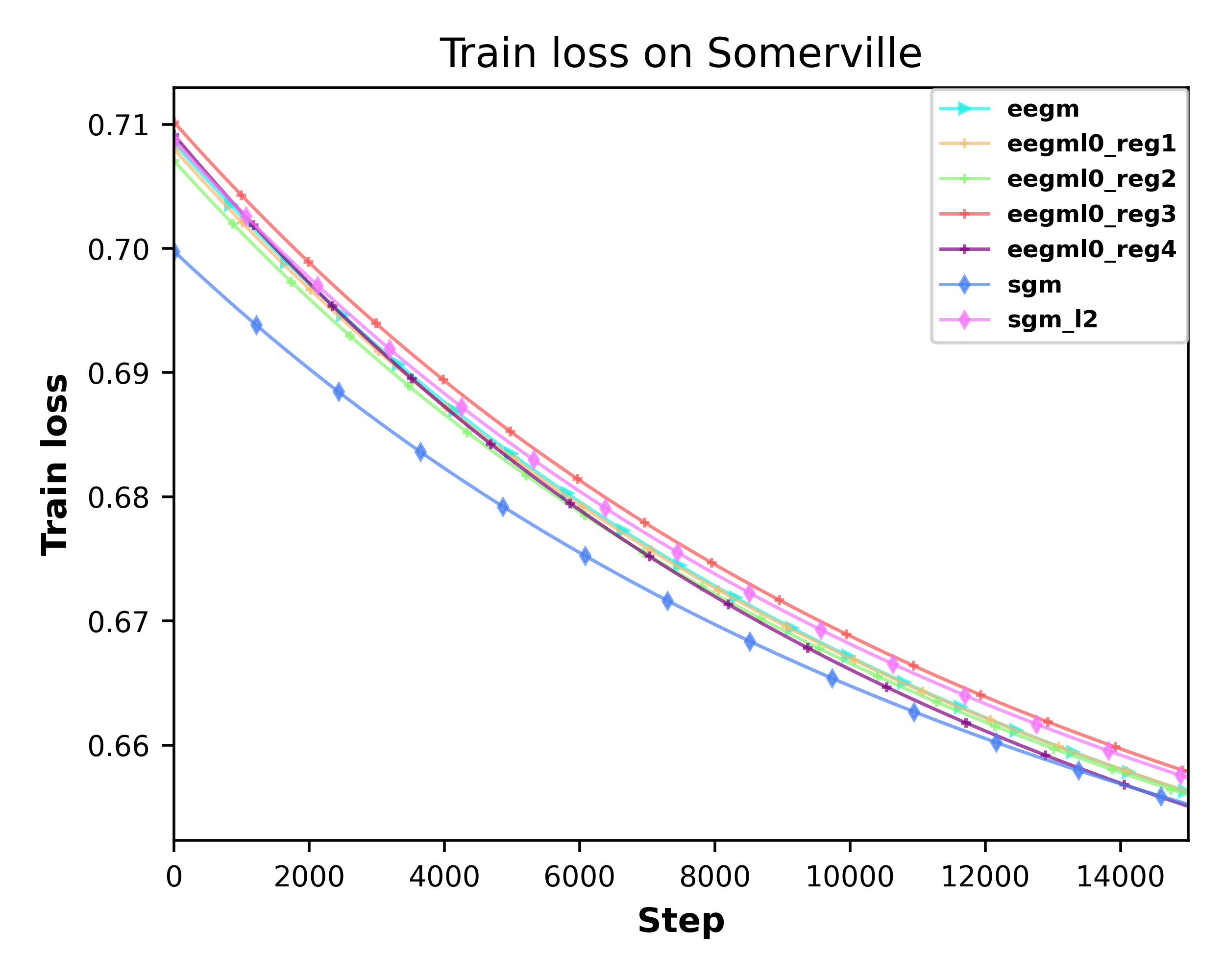}}  &
		\subfigure[Test loss]{\label{fig:Somerville_test_loss}\includegraphics[width=80mm]{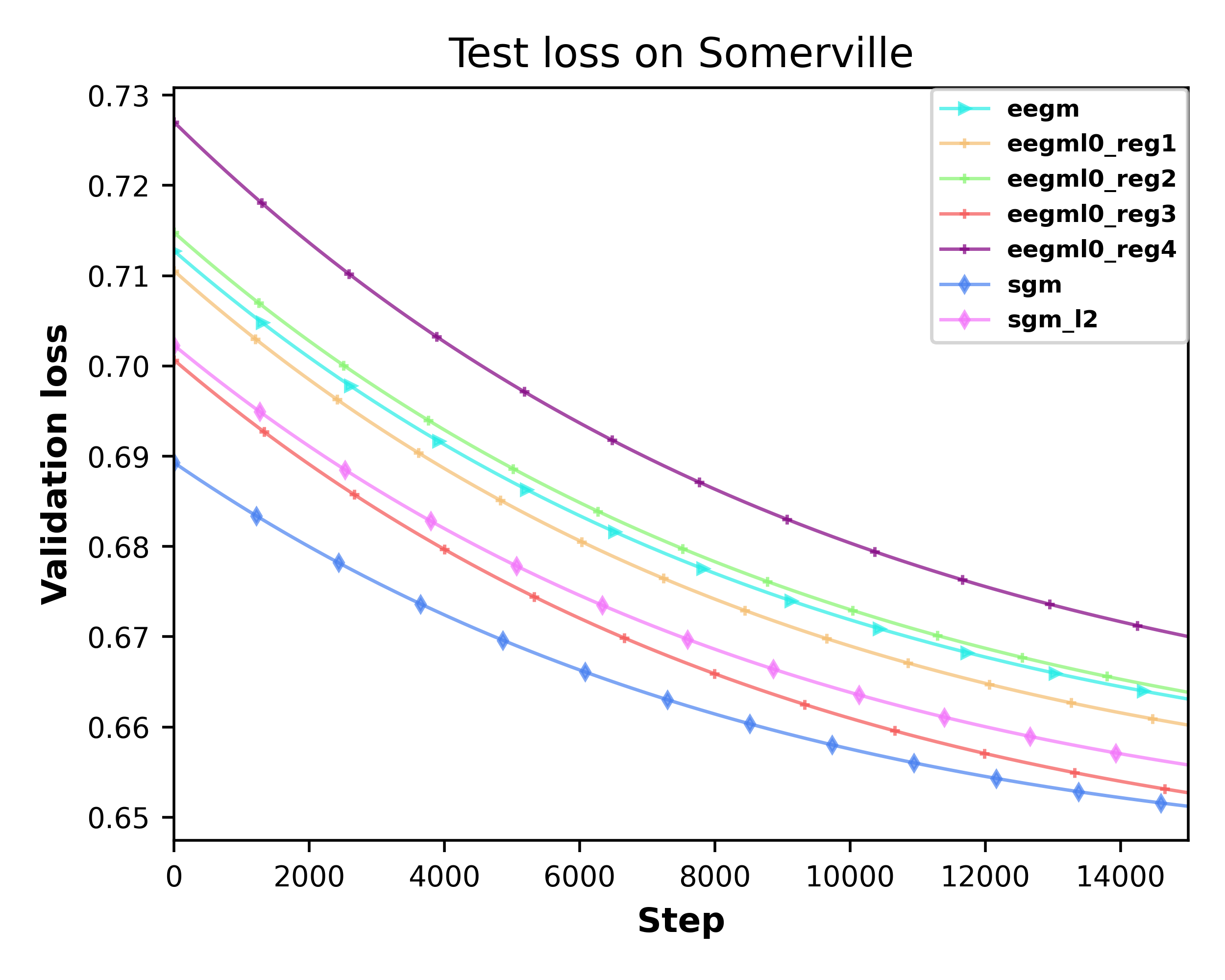}} 
	\end{tabular}
	\vspace{-.2cm}
	\caption{Somerville Happiness Survey.}
	\vspace{-.2cm}
	\label{fig:Somerville}
\end{figure*}

\begin{figure*}[h!]
	\centering    
	\begin{tabular}{c c}		
		\subfigure[Train loss]{\label{fig:PIMA_train_loss}\includegraphics[width=80mm]{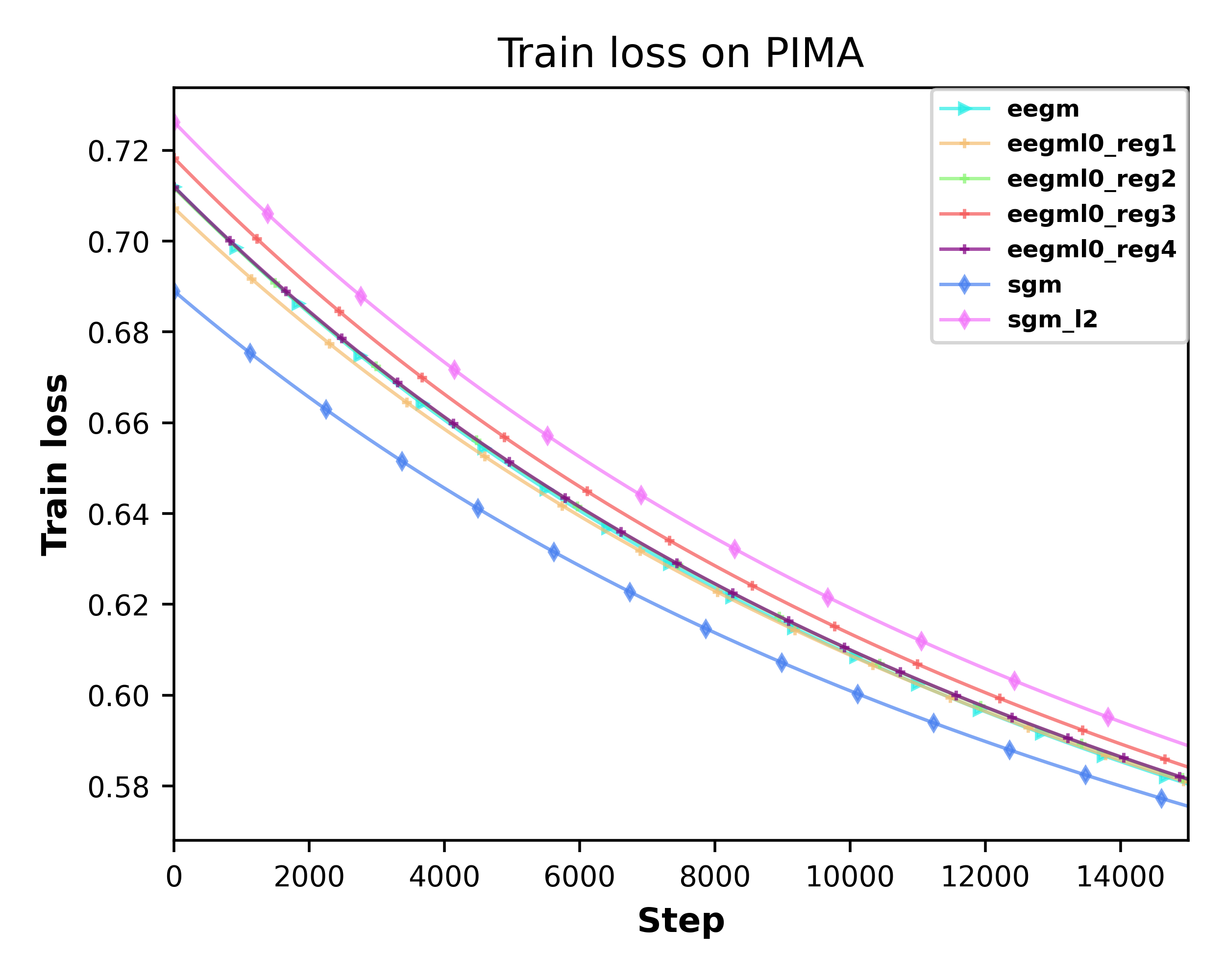}}  &
		\subfigure[Test loss]{\label{fig:PIMA_test_loss}\includegraphics[width=80mm]{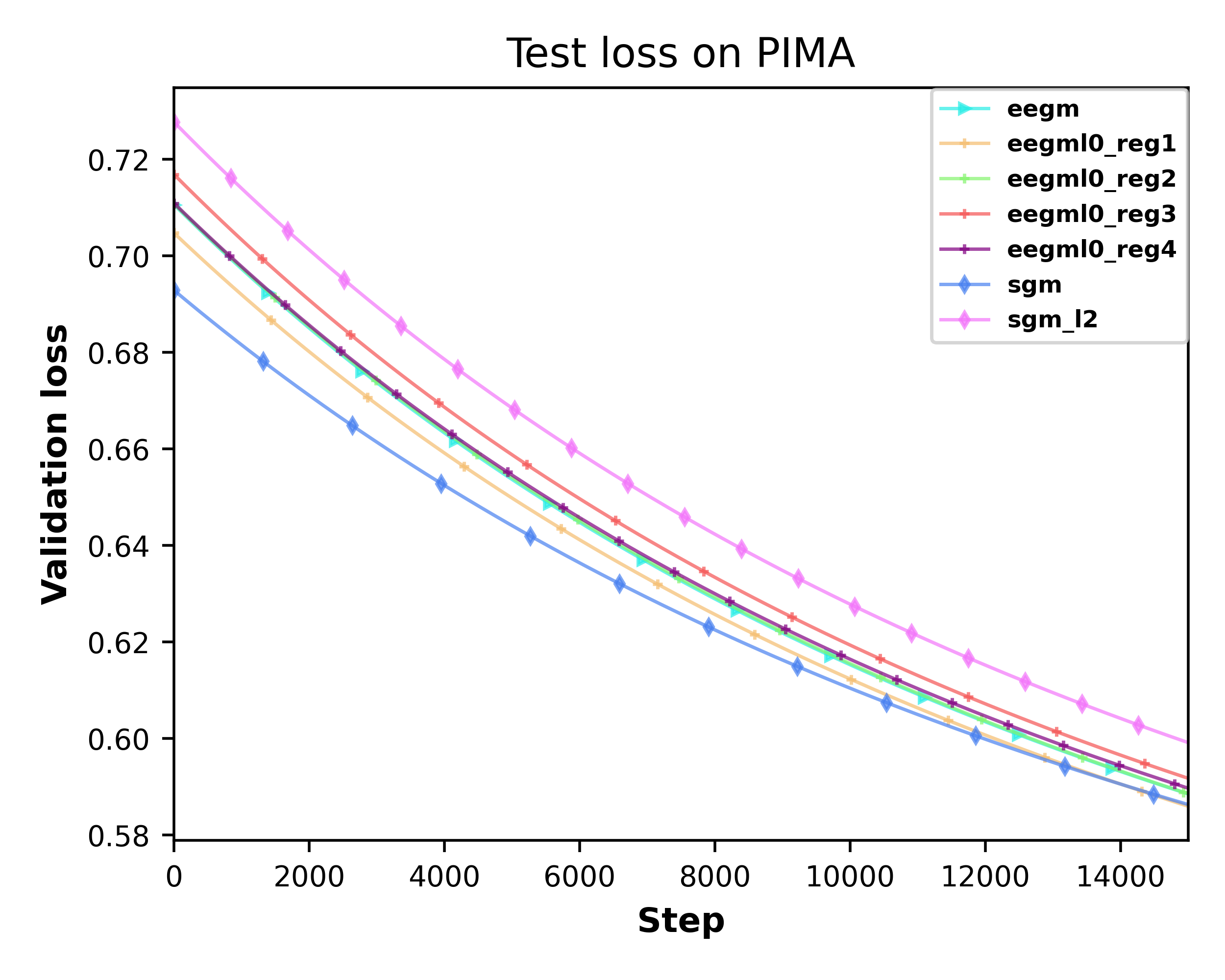}} 
	\end{tabular}
	\vspace{-.2cm}
	\caption{PIMA.}
	\vspace{-.2cm}
	\label{fig:PIMA}
\end{figure*}

\begin{figure*}[h!]
	\centering    
	\begin{tabular}{c c}		
		\subfigure[Train loss]{\label{fig:Ionosphere_train_loss}\includegraphics[width=80mm]{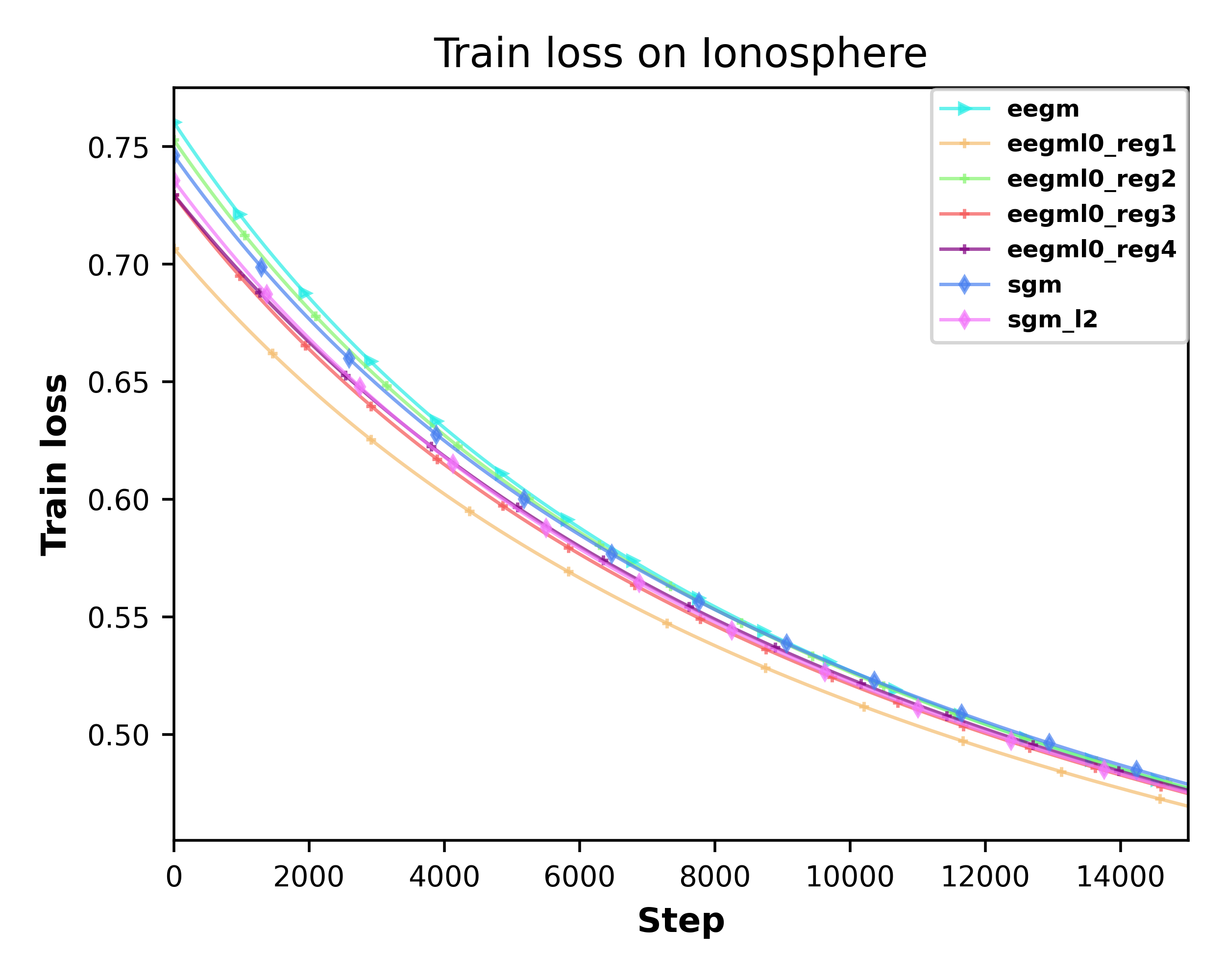}}  &
		\subfigure[Test loss]{\label{fig:Ionosphere_test_loss}\includegraphics[width=80mm]{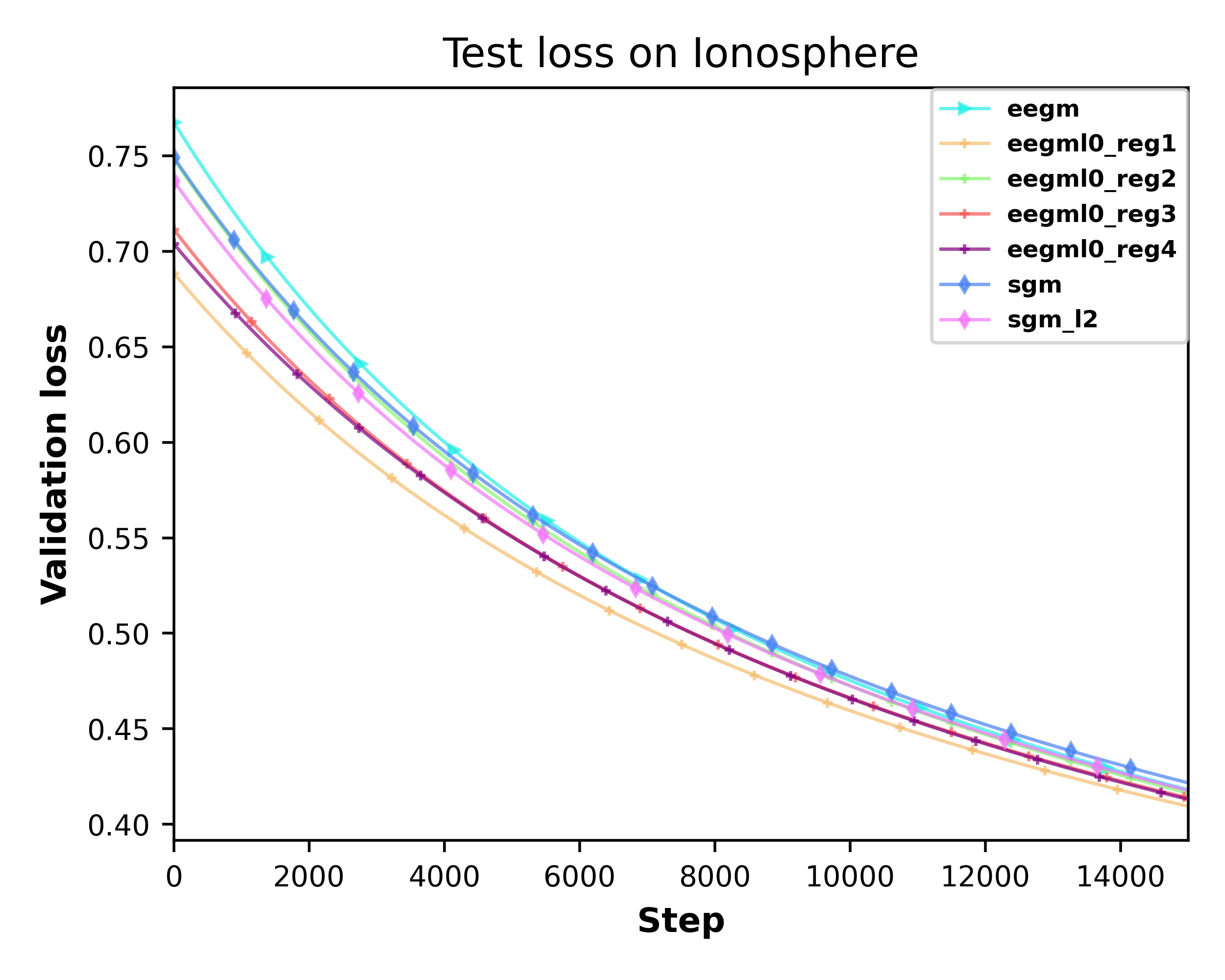}} 
	\end{tabular}
	\vspace{-.2cm}
	\caption{Ionosphere.}
	\vspace{-.2cm}
	\label{fig:Ionosphere}
\end{figure*}

\begin{figure*}[h!]
	\centering    
	\begin{tabular}{c c}		
		\subfigure[Train loss]{\label{fig:Gun_Point_train_loss}\includegraphics[width=80mm]{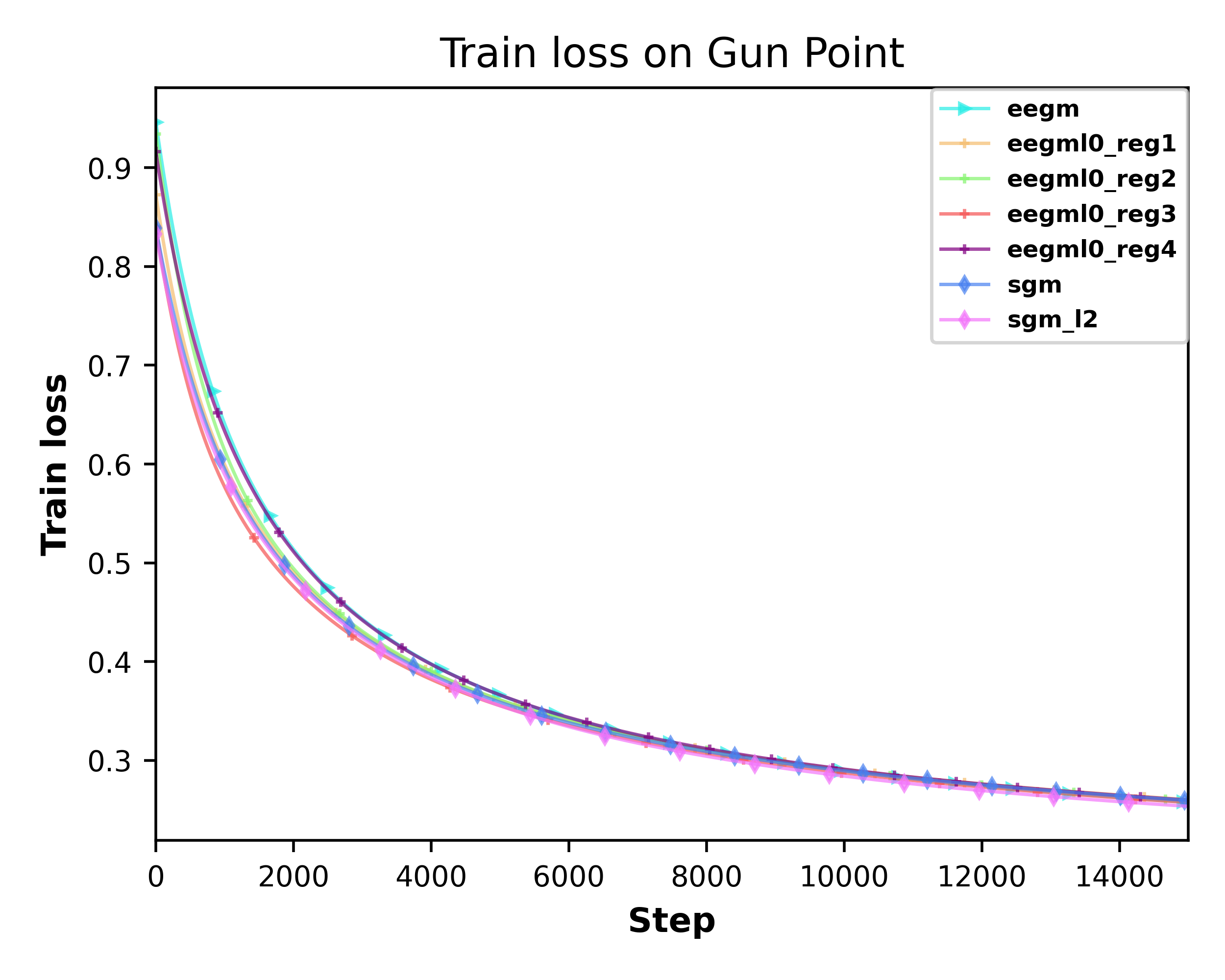}}  &
		\subfigure[Test loss]{\label{fig:Gun_Point_test_loss}\includegraphics[width=80mm]{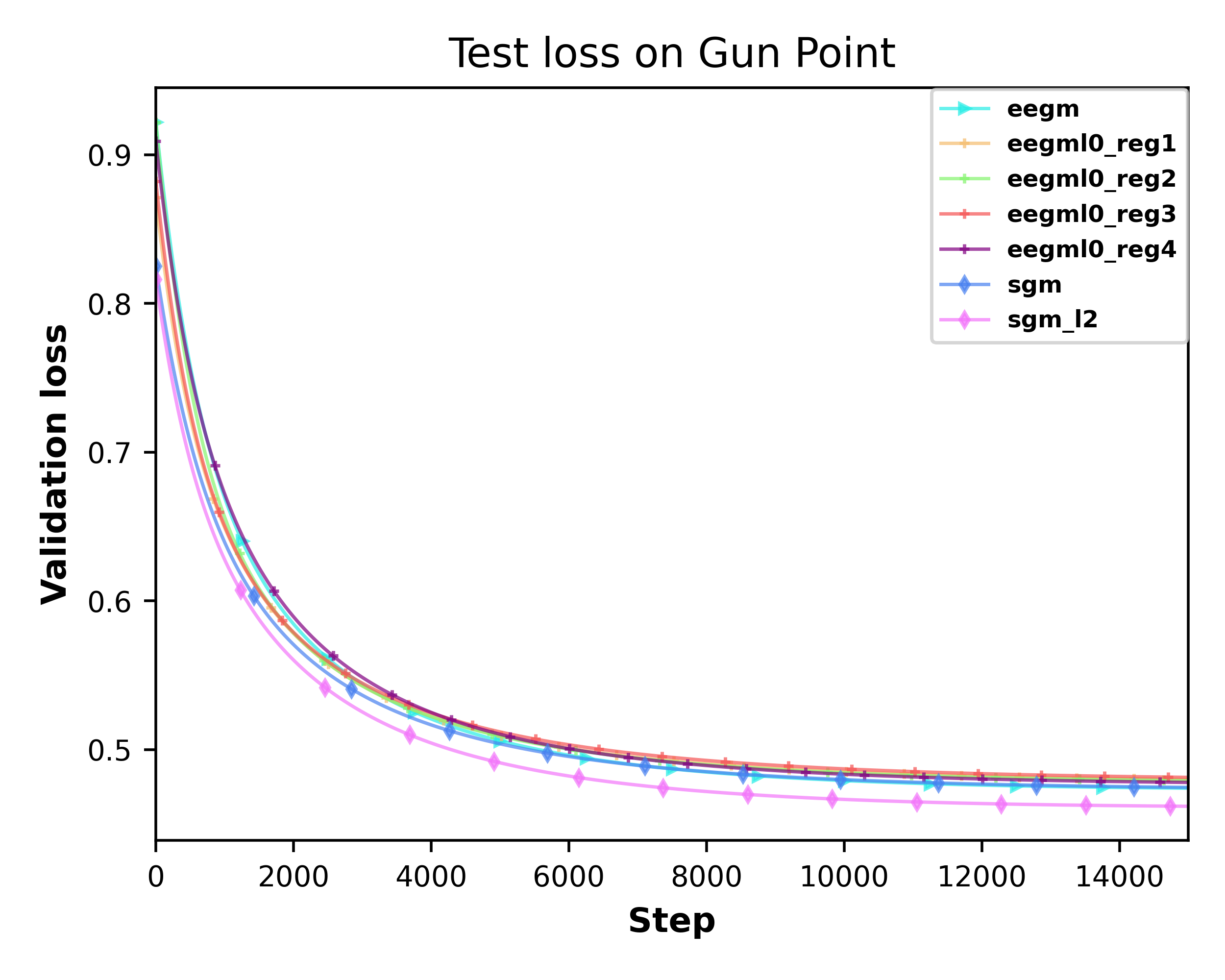}} 
	\end{tabular}
	\vspace{-.2cm}
	\caption{Gunpoint.}
	\vspace{-.2cm}
	\label{fig:Gun_Point}
\end{figure*}

\begin{figure*}[h!]
	\centering    
	\begin{tabular}{c c}		
		\subfigure[Train loss]{\label{fig:Coffee_train_loss}\includegraphics[width=80mm]{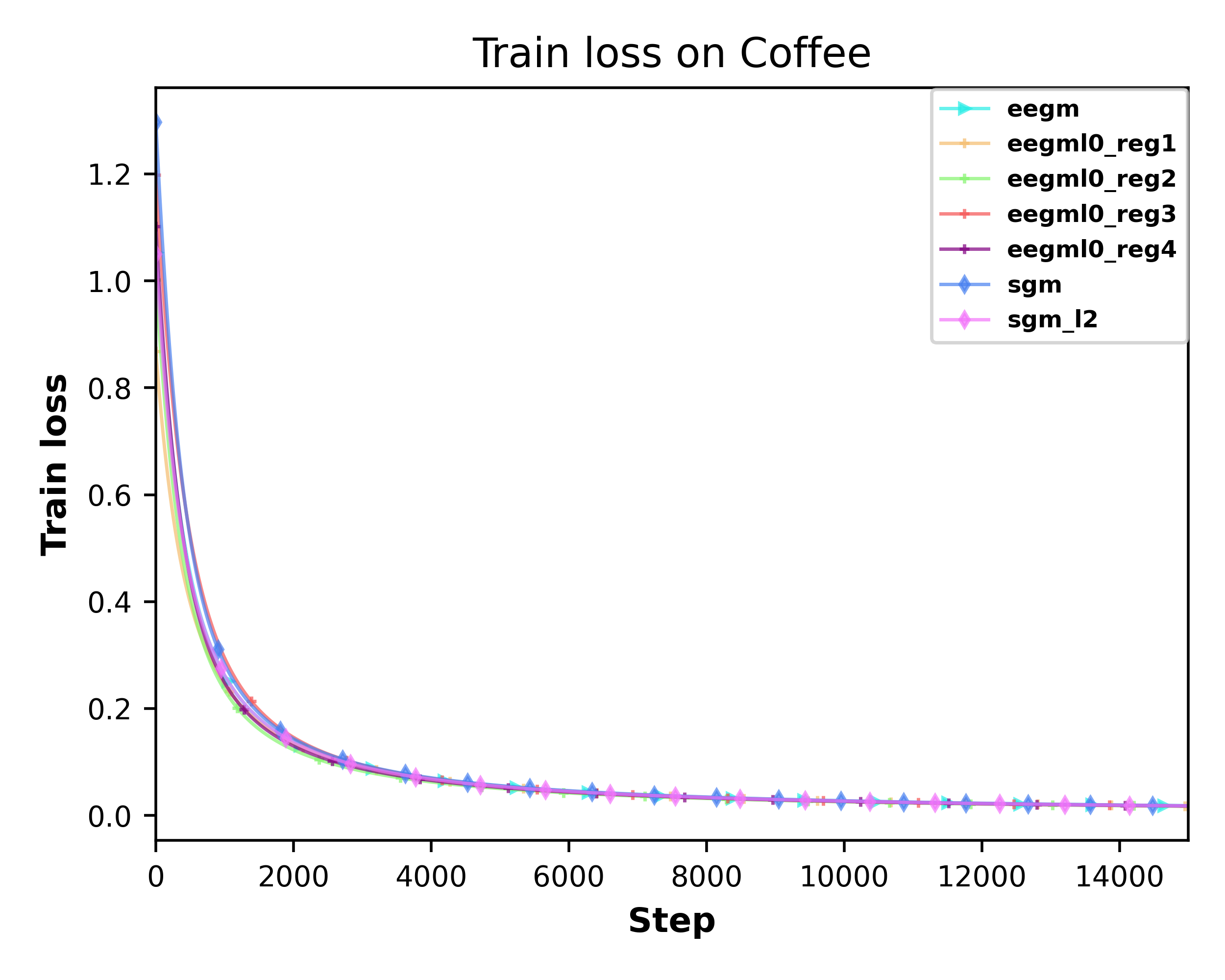}}  &
		\subfigure[Test loss]{\label{fig:Coffee_test_loss}\includegraphics[width=80mm]{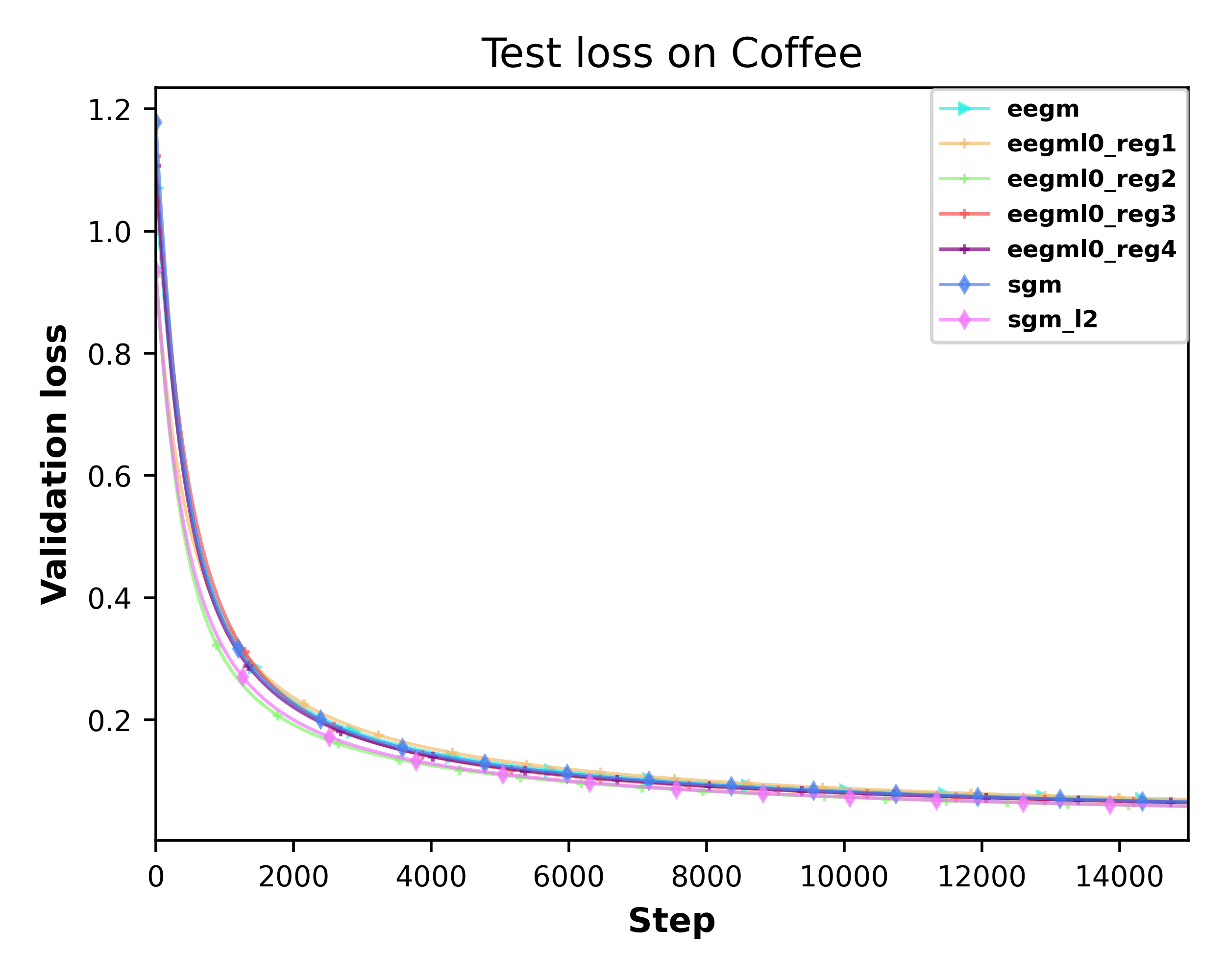}} 
	\end{tabular}
	\vspace{-.2cm}
	\caption{Coffee.}
	\vspace{-.2cm}
	\label{fig:Coffee}
\end{figure*}

\begin{figure*}[h!]
	\centering    
	\begin{tabular}{c c}		
		\subfigure[Spect Heart]{\label{fig:Spect_Heart_accuracy}\includegraphics[width=80mm]{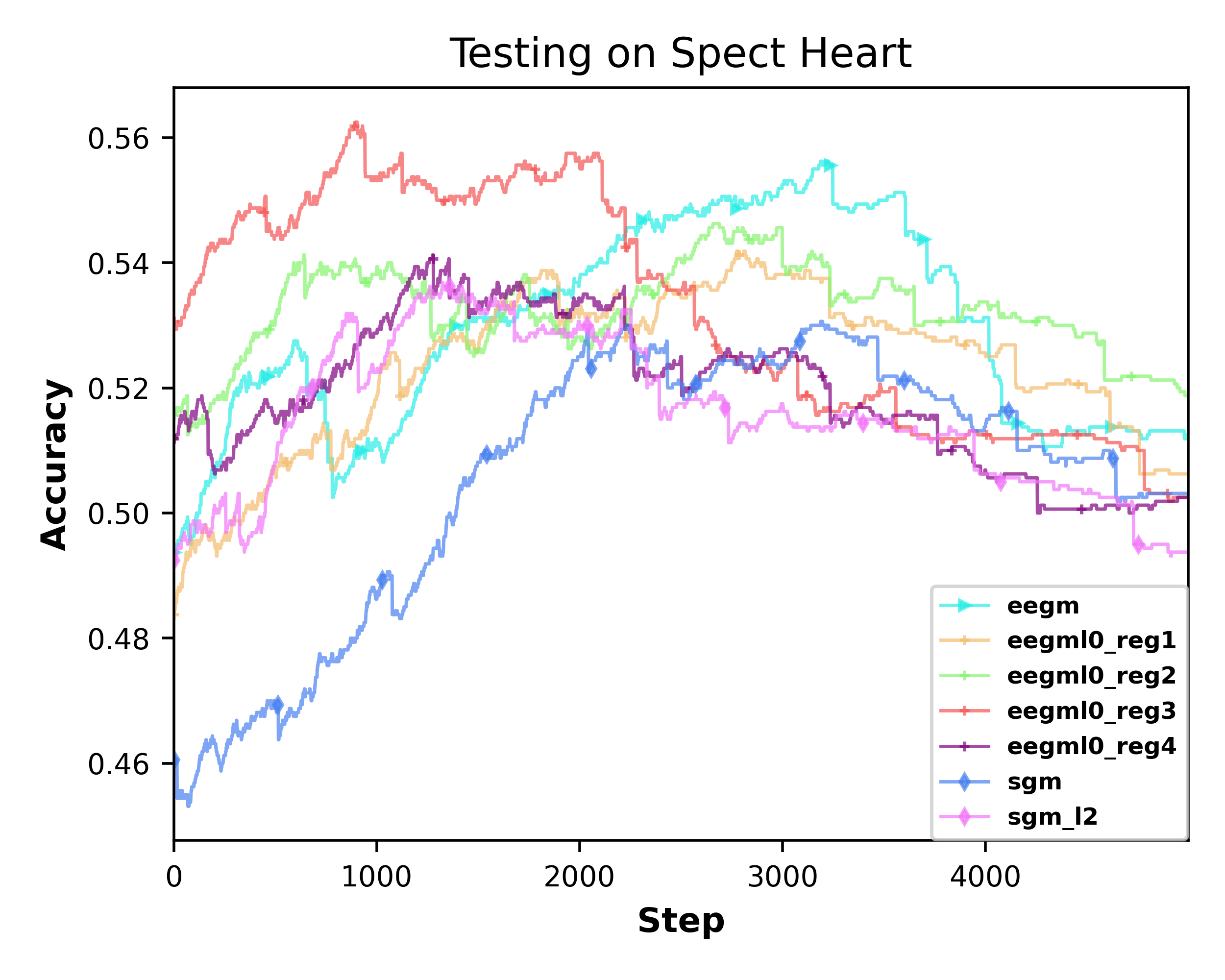}}  &
		\subfigure[Connectionist Bench]{\label{fig:Connectionist_Bench_accuracy}\includegraphics[width=80mm]{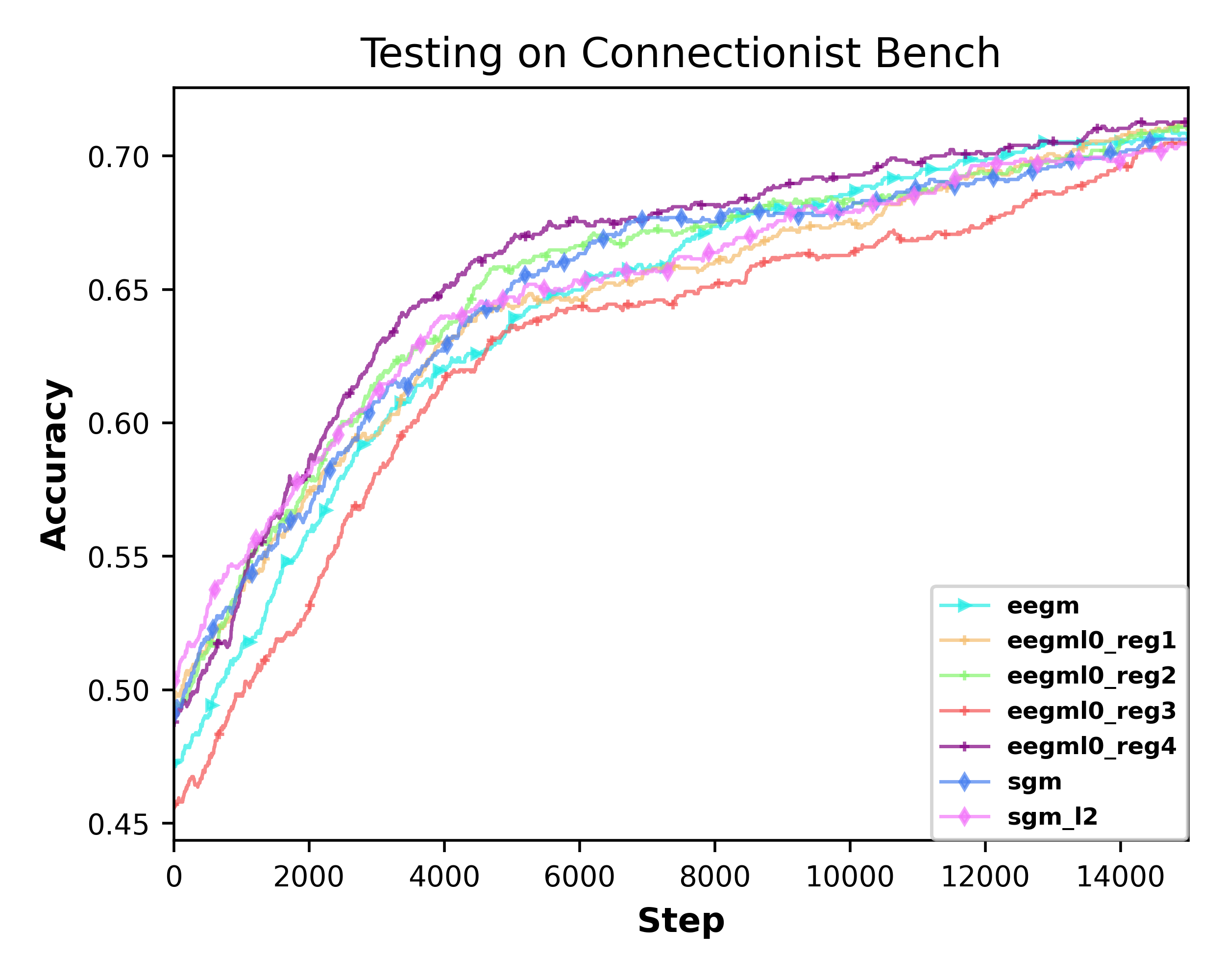}} 
	\end{tabular}
	\vspace{-.2cm}
	\caption{Accuracy: Spect Heart and Connectionist Bench.}
	\vspace{-.2cm}
	\label{fig:Spect_Heart_Connectionist_Bench}
\end{figure*}

\begin{figure*}[h!]
	\centering    
	\begin{tabular}{c c}		
		\subfigure[Divorce]{\label{fig:Divorce_accuracy}\includegraphics[width=80mm]{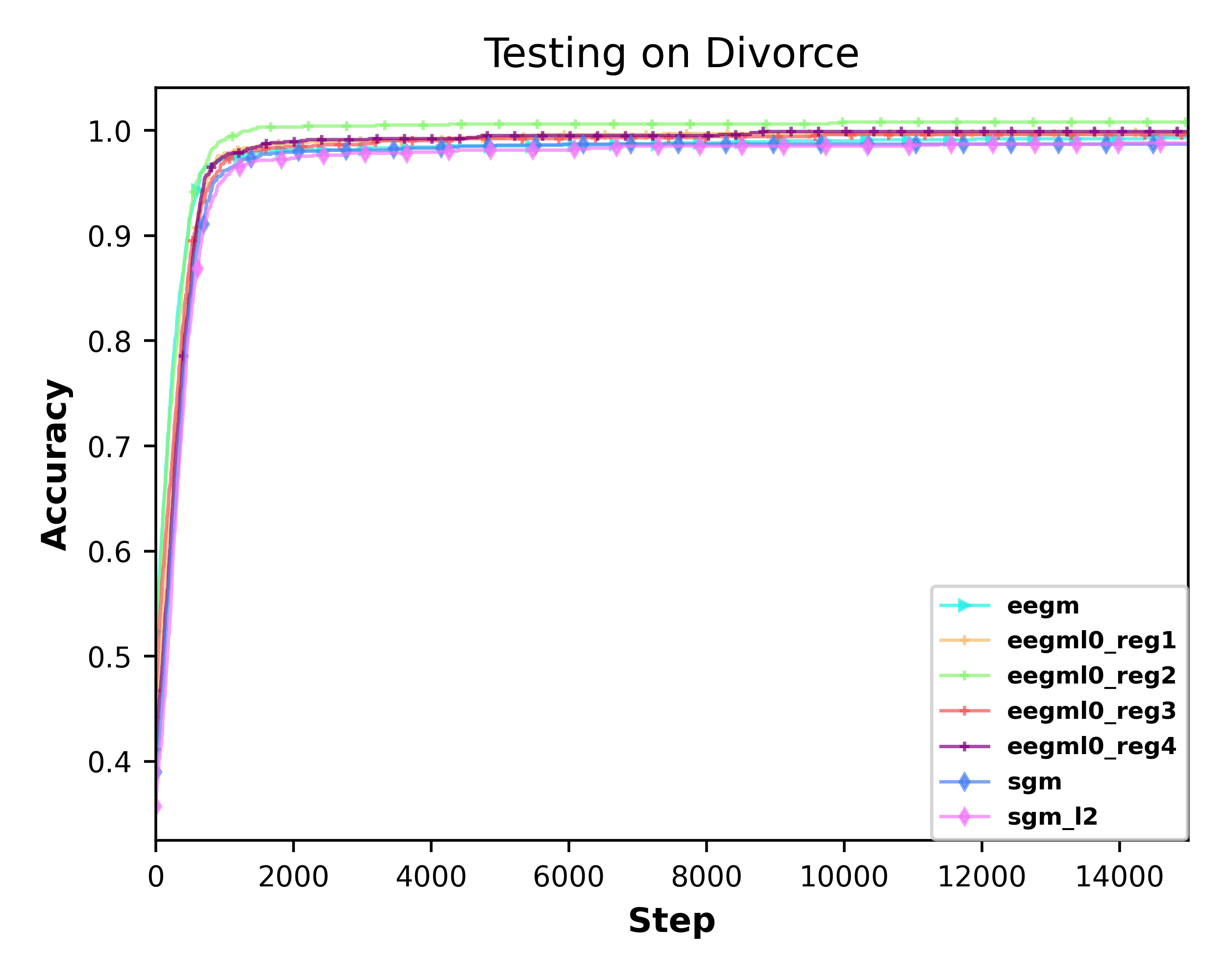}}  &
		\subfigure[Breast Cancer]{\label{fig:Breast_Cancer_accuracy}\includegraphics[width=80mm]{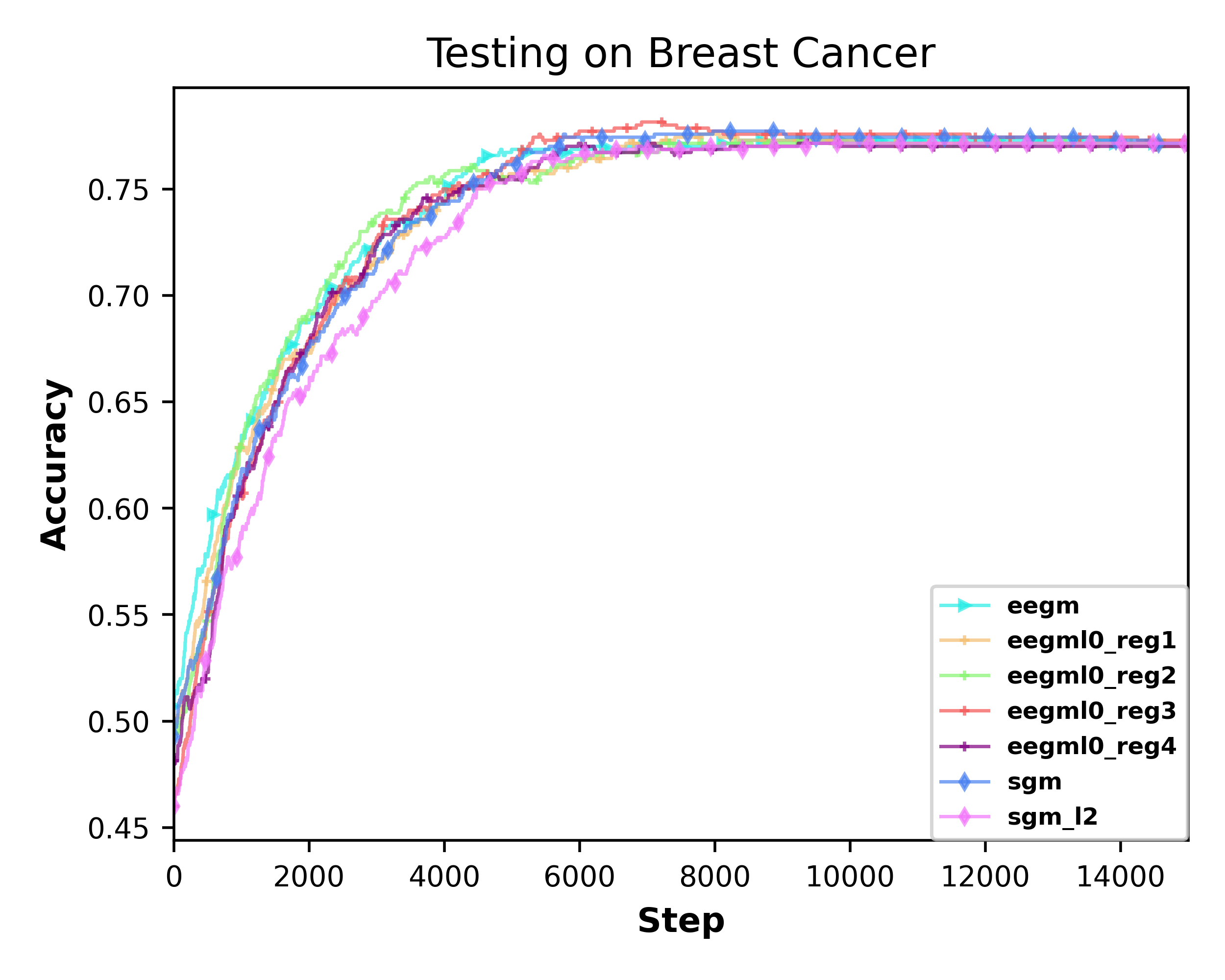}} 
	\end{tabular}
	\vspace{-.2cm}
	\caption{Accuracy: Divorce and Breast Cancer.}
	\vspace{-.2cm}
	\label{fig:Accuracy_Divorce_Breast_Cancer}
\end{figure*}

\begin{figure*}[h!]
	\centering    
	\begin{tabular}{c c}		
		\subfigure[Somerville]{\label{fig:Somerville_accuracy}\includegraphics[width=80mm]{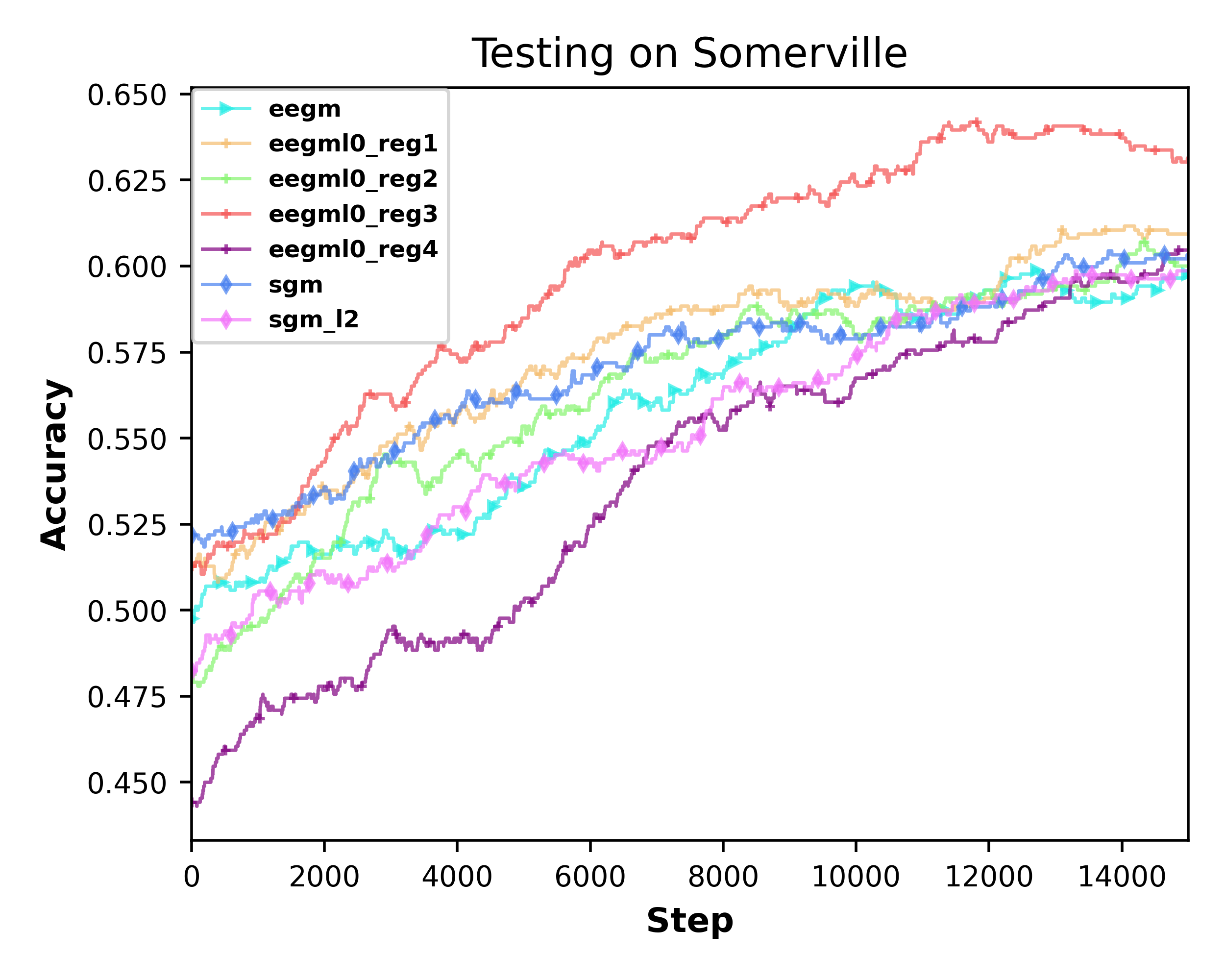}}  &
		\subfigure[PIMA]{\label{fig:PIMA_accuracy}\includegraphics[width=80mm]{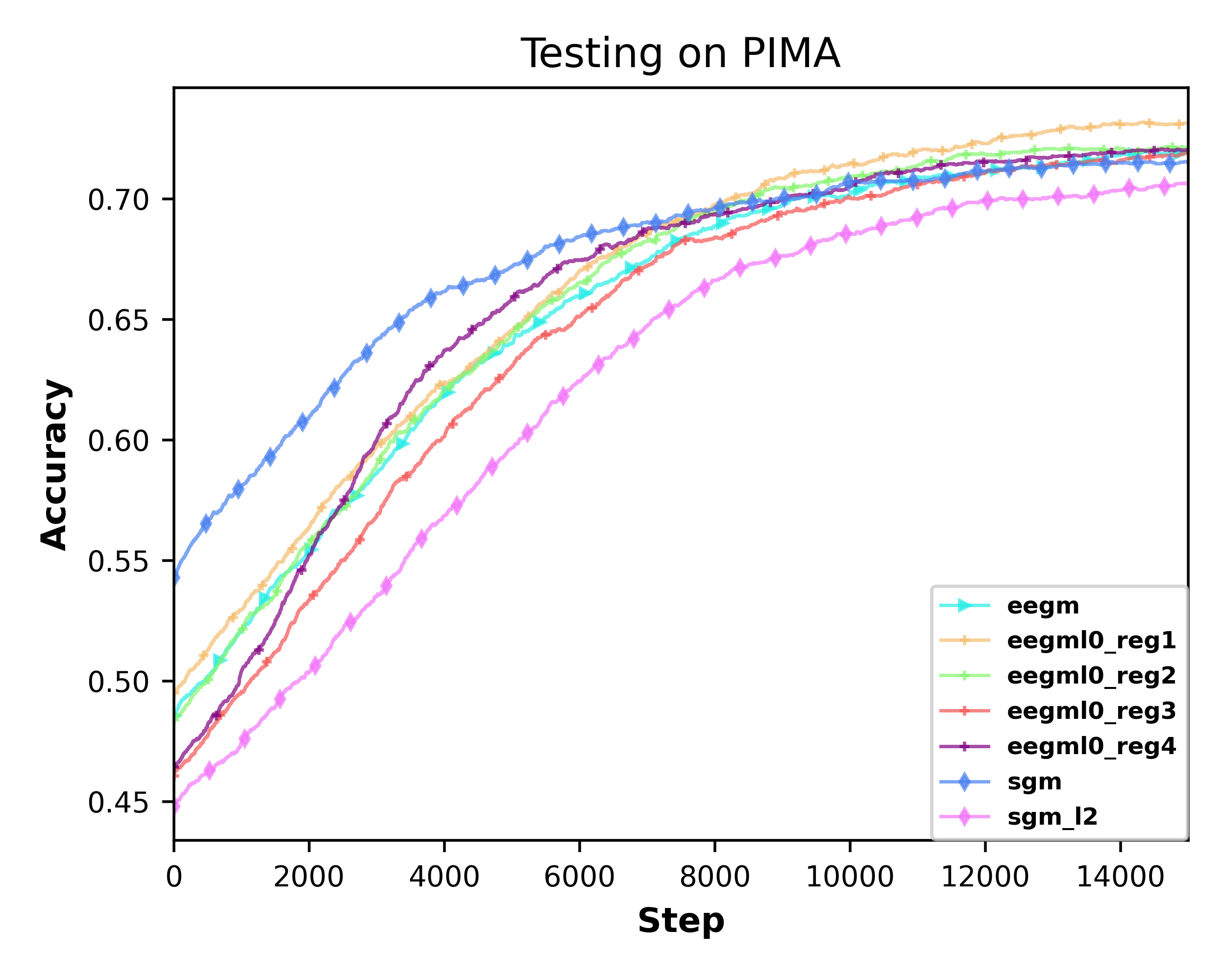}} 
	\end{tabular}
	\vspace{-.2cm}
	\caption{Accuracy: Somerville and PIMA.}
	\vspace{-.2cm}
	\label{fig:Accuracy_Somerville_PIMA}
\end{figure*}

\begin{figure*}[h!]
	\centering    
	\begin{tabular}{c c}		
		\subfigure[Ionosphere]{\label{fig:Ionosphere_accuracy}\includegraphics[width=80mm]{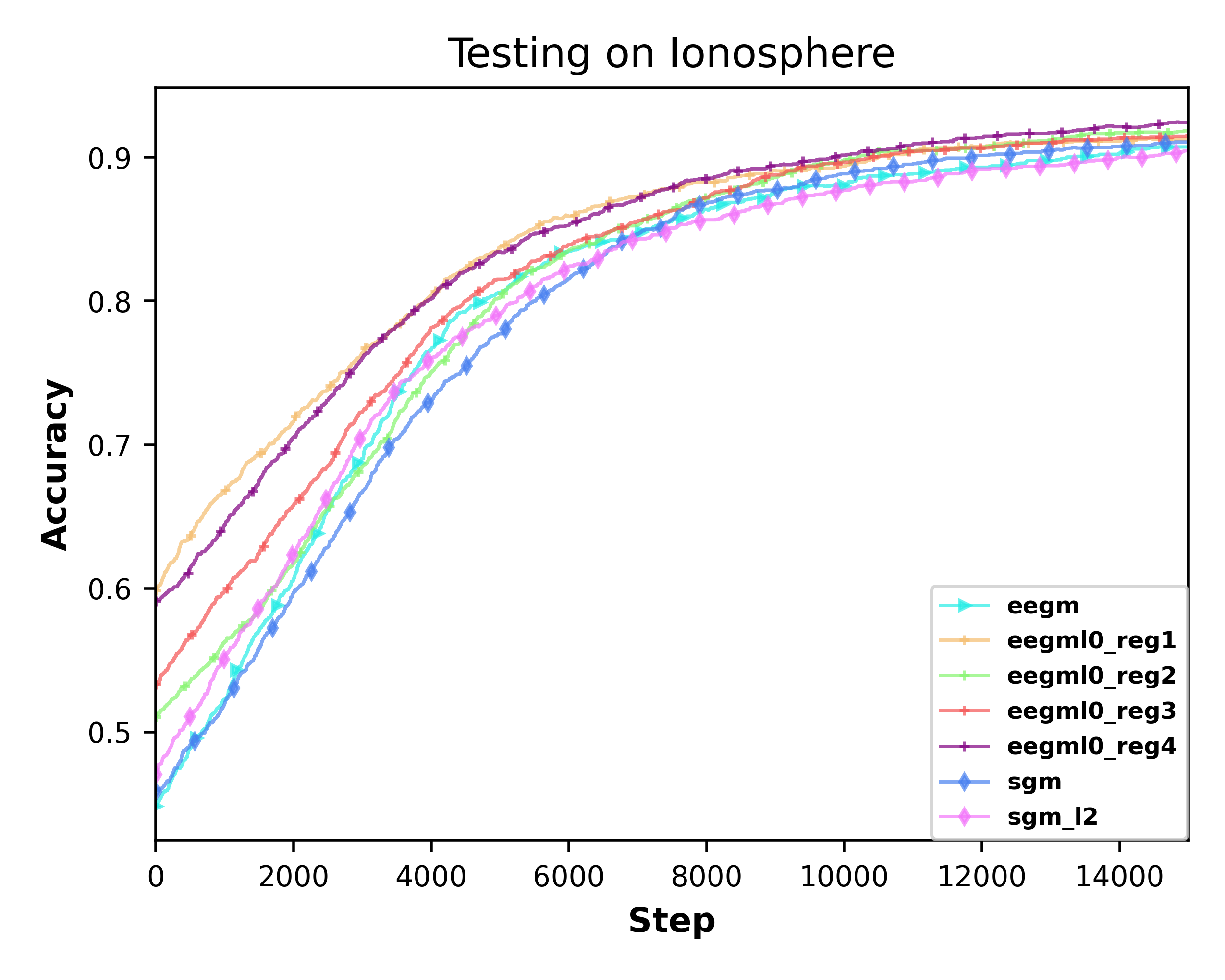}}  &
		\subfigure[Gunpoint]{\label{fig:Gun_Point_accuracy}\includegraphics[width=80mm]{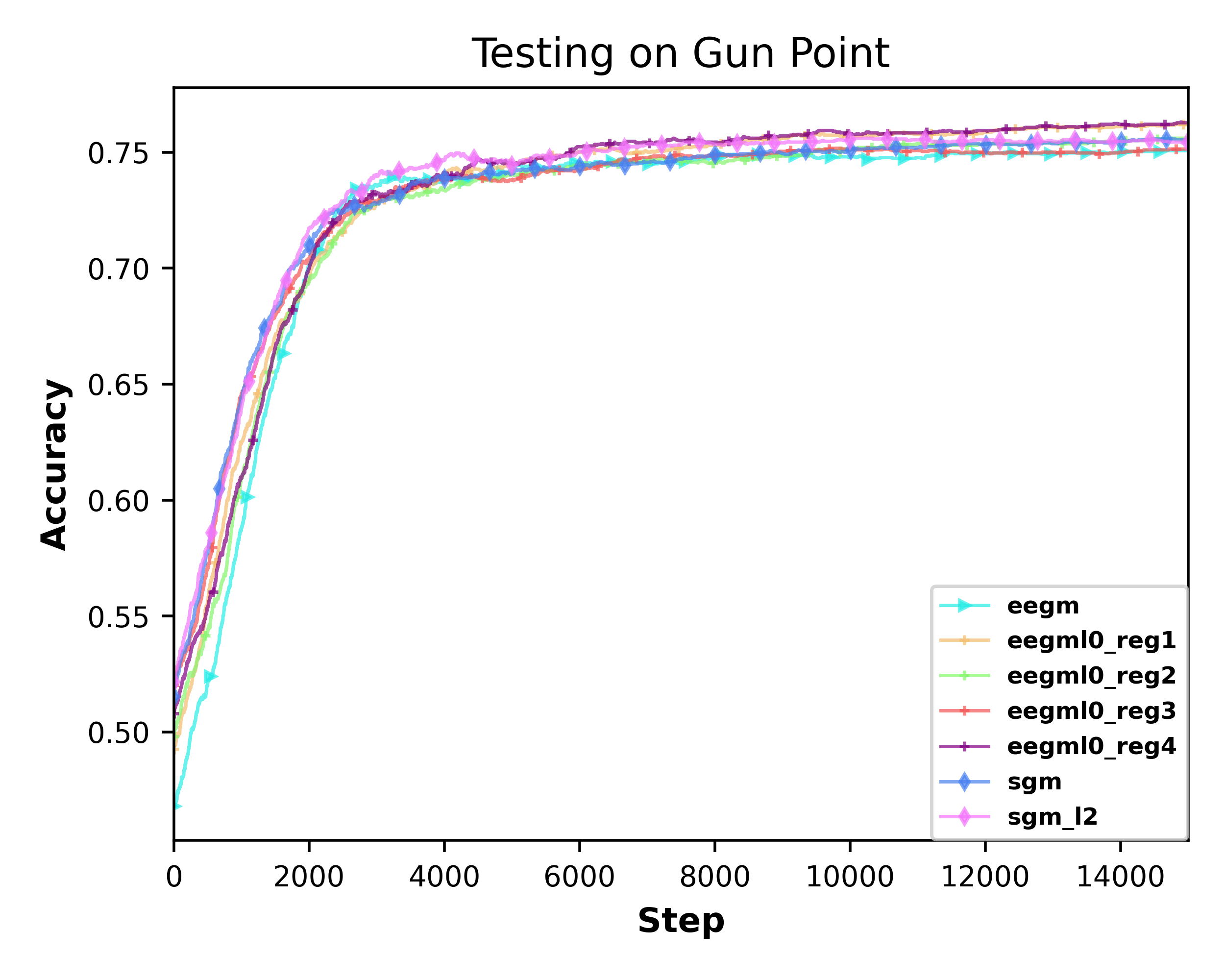}} 
	\end{tabular}
	\vspace{-.2cm}
	\caption{Accuracy: Ionosphere and Gunpoint.}
	\vspace{-.2cm}
	\label{fig:Accuracy_Ionosphere_Gun_Point}
\end{figure*}

\begin{figure}[h!]
	\centering
	\includegraphics[width=80mm]{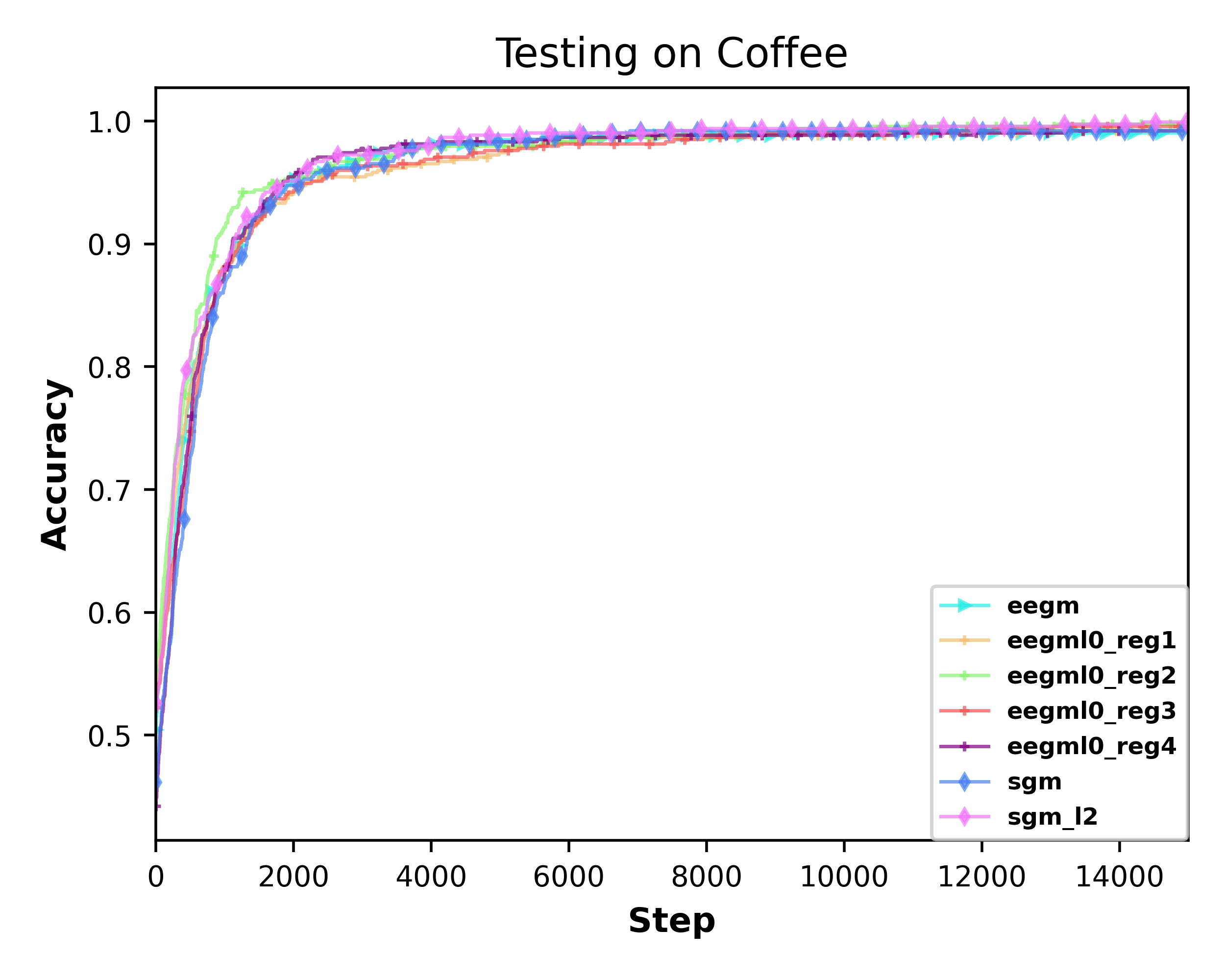}
	\caption{Accuracy: Coffee.} 
	\label{fig:Accuracy_Coffee}
\end{figure}

\revised{Table \ref{tab:comparison_table} shows the average test accuracy of 20 trials of all experiments. 
All experiments use the original SGD optimizer to optimize the neural network model. Training was set to run for 15,000 epochs in almost all experiments. For experiments in the Spect Heart dataset, the model is trained with only 5,000 epochs to avoid overfitting. It is evident that the proposed algorithm EEGML0 outperforms the other methods in all the nine datasets. For instance, by Spect Heart, EEGML0 + reg 3 gets 56.25\% as the accuracy, which is better compared to 53.93\% and 54.06\%, the corresponding score obtained by EEGM and SGM. Especially, by the Somerville happiness survey, the proposed algorithm gets the most distinct result, improving the accuracy from 59.88\% to 64.18\%. Moreover, it obtains a perfect prediction, \ie 100\% by the Divorce predictors dataset. Similarly, on the other datasets, EEGML0 gains a superior accuracy. 
Altogether, it is observed that the EEGML0 algorithm with different regularizer options is better compared to the EEGM approach with respect to the prediction accuracy by most of the considered datasets.}

\revised{An additional analysis was conducted on the training curve of each regularizer to compare the performance between the regularizer options. The curves together with the accuracy for all experiments for the first three datasets, \ie Spect Heart, and Connectionist Bench, and Divorce Predictor are sketched in Figures~\ref{fig:Spect_Heart},\ref{fig:Connectionist_Bench},\ref{fig:Divorce}. Only on the Spect Heart dataset, with respect to the train loss, it is clear that EEGML0 + reg3 accounts for the smallest loss compared to the other configurations. Meanwhile, by the two remaining datasets, there is no big difference in the loss. Similarly, as shown in Figures~\ref{fig:Breast_Cancer},\ref{fig:Somerville},\ref{fig:PIMA}, we witness the same trend for the Breast Cancer, Somerville, and PIMA datasets. In Figures~\ref{fig:Ionosphere},\ref{fig:Gun_Point},\ref{fig:Coffee}, the train loss and test loss for the Ionosphere, Gun Point, and Coffee datasets are shown, and it is evident that also in these cases the difference in the loss of the models is marginal. Concerning validation loss, EEGML0 + reg3 is again the one that yields the smallest loss, among others. From the aforementioned figures, it is observed that the EEGML0 algorithm with regularizer 3 has the best generalization performance since it always has the smallest gap between train loss and validation loss. In terms of accuracy, regularizer 3 also gives the best average test accuracy on a major number of datasets including Spect Heart, Connectionist bench, Divorce predictors, and Somerville happiness survey.} 


The accuracy for Spect Heart and Connectionist Bench is 
shown in Figure~\ref{fig:Spect_Heart_Connectionist_Bench}. \revised{It is evident that EEGML0 + reg 3 yields the best accuracy by the datasets. However, EEGML0 + reg1 is the configuration that obtains the best accuracy on Connectionist Bench. Concerning PIMA, EEGML0 + reg 1 is the one that gives the best accuracy. By the Divorce Predictor and Breast Cancer Coimbra datasets, as shown in Figure~\ref{fig:Accuracy_Divorce_Breast_Cancer}, EEGML0 + reg2 gets a slightly better accuracy compared to other models.} By the remaining datasets 
the performance is almost the same for all the configurations. Figure~\ref{fig:Accuracy_Somerville_PIMA} depicts the results obtained for Sommerville Happiness Survey and PIMA, and it can be seen that there is no distinct difference among the results of these configurations.

\revised{Finally, Figure~\ref{fig:Accuracy_Ionosphere_Gun_Point} and Figure~\ref{fig:Accuracy_Coffee} depict the experimental results for the last three datasets, \ie Ionosphere, GunPoint, and Coffee. Overall, on these datasets, EEGML0 + reg1 achieves the best performance by Ionosphere dataset, and there is no distinct difference in accuracy by the remaining datasets. Altogether, the conclusion is that depending on the input data, EEGML0 is usually the model that gets the best accuracy by most of the datasets.}


    
    \textbf{Threats to validity.} 
	Threats to {\em internal validity} concern the confounding factors in the approach and evaluation that could have influenced the final results. The comparison with the baseline might be susceptible to bias. The threat was mitigated by following the same experimental settings that have been used in the original paper. Since the source code of the original work \cite{xiong_convergence_2020} is not available, it is not possible to confirm their results. 
	The evaluation was executed several times to get the best results. Here, only the relative comparison with the results obtained by Xiong \etal \cite{xiong_convergence_2020} is given.
	Moreover, the experiments were ran for several times to make sure that the results are consistent across the trials. 
	
	The main threat to {\em external validity} is related to the generalizability of the findings. A probable threat is that the datasets might not fully simulate a real-world setting as only benchmark datasets were used for testing. 
	To tackle the issues, various datasets coming from different domains were used in the evaluation. However, considering data from other sources can help further eliminate the threat, and this can be considered as future work.

	\section{Conclusion and future work} \label{sec:Conclusions}
	
	This paper introduced a novel entropy function with smoothing $l_0$ regularization for feed-forward neural networks. An empirical evaluation using real-world datasets demonstrated that the newly proposed approach enables feed-forward neural networks to improve the prediction performance. Moreover, the algorithm also outperforms well-established baselines. In this respect, the conceived algorithm is expected to help neural networks increase their efficiency and effectiveness.
	
	\revised{For future work, the plan is to evaluate the proposed $h$-regularization approach with larger as well as diverse datasets. Moreover, it is necessary to consider experimenting with the conceived algorithm on data coming from different application domains, including Computer Vision, and Natural Language Processing, among others. The $h$-regularization method is supposed to come in handy in training convolutional neural networks (CNNs), which have been developed for image recognition and classification.}

	\section*{Declarations}

	\vspace{.2cm}
	\noindent
	\textbf{Competing interest.} The authors declare no conflict of interest.

	\vspace{.2cm}
	\noindent
	\textbf{Author contributions.} Trong-Tuan Nguyen implements the algorithm, and runs the evaluation; Van-Dat Thang implements the algorithm; Nguyen Van Thin conceptualizes the approach, performs the mathematical proofs, writes the paper; Phuong T. Nguyen supervises the whole work, and contributes in the writing of the manuscript.

	
	\vspace{.2cm}
	\noindent
	\textbf{Ethical approval.} This article does not contain any studies with human  participants or animals performed by any of the authors.

	\vspace{.2cm}
	\noindent
	\textbf{Data availability.} All the authors declare that they have no data to make available for this submission. 

\end{document}